\documentclass[10pt,twocolumn,letterpaper]{article}

\usepackage{cvpr}
\usepackage{times}
\usepackage{epsfig}
\usepackage{graphicx}
\usepackage{amsmath}
\usepackage{amssymb}
\usepackage{booktabs}

\usepackage{color}
\usepackage{dsfont}
\usepackage[utf8]{inputenc}
\usepackage[percent]{overpic}
\usepackage[normalem]{ulem}
\usepackage{textcomp}
\usepackage{multirow}

\usepackage{snapshot}           

\usepackage[pagebackref=true,breaklinks=true,letterpaper=true,colorlinks,bookmarks=false]{hyperref}

\newcommand{\comment}[1]{}

\newcommand{\bmu}{\boldsymbol{\mu}}

\newcommand{\bI}{\mathbf{I}}

\newcommand{\bo}{\mathbf{o}}
\newcommand{\bw}{\mathbf{w}}
\newcommand{\bp}{\mathbf{p}}

\newcommand{\bK}{\mathbf{K}}
\newcommand{\bsigma}{\boldsymbol{\sigma}}
\newcommand{\bSigma}{\boldsymbol{\Sigma}}

\newcommand{\bX}{\mathbf{X}}
\newcommand{\bM}{\mathbf{M}}
\newcommand{\bU}{\mathbf{U}}

\newcommand{\bx}[0]{\mathbf{x}}
\newcommand{\bcorr}[0]{\mathbf{q}}
\newcommand{\by}[0]{\mathbf{y}}

\newcommand{\IR}{\mathds{R}}

\newcommand{\diag}{\operatorname{diag}}

\newcommand{\vectorize}{\operatorname{Vec}}

\newcommand{\fig}[1]{Fig.~\ref{fig:#1}}
\newcommand{\figs}[2]{Figs.~\ref{fig:#1} and~\ref{fig:#2}}
\newcommand{\tbl}[1]{Table~\ref{tbl:#1}}
\newcommand{\secref}[1]{Section~\ref{sec:#1}}
\newcommand{\refsec}[1]{Section~\ref{sec:#1}}

\newcommand{\loss}{\mathcal{L}}

\definecolor{orange}{rgb}{1,0.5,0}
\definecolor{blue}{rgb}{0,0,0.6}

\definecolor{color1}{RGB}{0,199,1}
\definecolor{color2}{RGB}{224,43,28}

\newcommand{\eq}[1]{Eq.~\eqref{eq:#1}}

\definecolor{darkolivegreen}{rgb}{0.5, 0.7, 0.3}

\newcommand{\kmyirmk}[1]{{\color{orange}{\bf #1}}}
\newcommand{\kyi}[1]{{\color{orange}{#1}}}
\newcommand{\KYI}[1]{{\color{orange}{\bf #1}}}

\newcommand{\PF}[1]{{\color{blue}{\bf #1}}}

\newcommand{\MS}[1]{{\color{magenta}{\bf #1}}}

\newcommand{\vincentrmk}[1]{{{\color{darkolivegreen}{\bf #1}}}}

\newcommand{\eduardrmk}[1]{{\color{red}{\bf #1}}}
\newcommand{\EDUARD}[1]{{\color{red}{\bf #1}}}

\renewcommand{\kmyirmk}[1]{}
\renewcommand{\kyi}[1]{#1}
\renewcommand{\KYI}[1]{}

\renewcommand{\PF}[1]{}

\renewcommand{\MS}[1]{}

\renewcommand{\vincentrmk}[1]{}

\renewcommand{\eduardrmk}[1]{}
\renewcommand{\EDUARD}[1]{}

\newcommand{\bE}{\mathbf{E}}

\cvprfinalcopy 

\def\cvprPaperID{****} 

\ifcvprfinal\fi
\begin{document}

\title{Learning to Find Good Correspondences}

\author{Kwang Moo Yi\textsuperscript{1,}\thanks{%
    First two authors contributed equally. K.M. Yi was at EPFL during the
    development of this project. This work was partially supported by EU FP7
    project MAGELLAN under grant ICT-FP7-611526 and by systems supplied by
    Compute Canada.%
  } \quad Eduard Trulls \textsuperscript{2,}\footnotemark[1] \quad Yuki
  Ono\textsuperscript{3} \quad Vincent Lepetit\textsuperscript{4}
  \quad Mathieu Salzmann\textsuperscript{2} \quad Pascal Fua\textsuperscript{2}\\
  {\small \textsuperscript{1}Visual Computing Group, University of
    Victoria}\quad
  {\small \textsuperscript{2}Computer Vision Laboratory, \'{E}cole Polytechnique F\'{e}d\'{e}rale de Lausanne}\\
  {\small \textsuperscript{3}Sony Imaging Products \& Solutions Inc.}\quad
  {\small \textsuperscript{4}Institute for Computer Graphics and Vision, Graz University of Technology}\\
  {\tt\small kyi@uvic.ca, \{firstname.lastname\}@epfl.ch, yuki.ono@sony.com, lepetit@icg.tugraz.at}
}

\maketitle


\begin{abstract}
  We develop a deep architecture to learn to find good correspondences for
  wide-baseline stereo.  Given a set of putative sparse matches and the camera
  intrinsics, we train our network in an end-to-end fashion to label the
  correspondences as inliers or outliers, while simultaneously using them to
  recover the relative pose, as encoded by the essential matrix.  Our
  architecture is based on a multi-layer perceptron operating on pixel
  coordinates rather than directly on the image, and is thus simple and
  small. We introduce a novel normalization technique, called Context
  Normalization, which allows us to process each data point separately while
  embedding global information in it, and also makes the network invariant to
  the order of the correspondences.  Our experiments on multiple challenging
  datasets demonstrate that our method is able to drastically improve the state
  of the art with little training data.
\end{abstract}


\vspace{-1em}

\section{Introduction}

\begin{figure}
\centering
\footnotesize
\begin{tabular}{@{}c@{}c@{}}
  \includegraphics[width=0.495\linewidth, trim=0 10 0 0, clip]{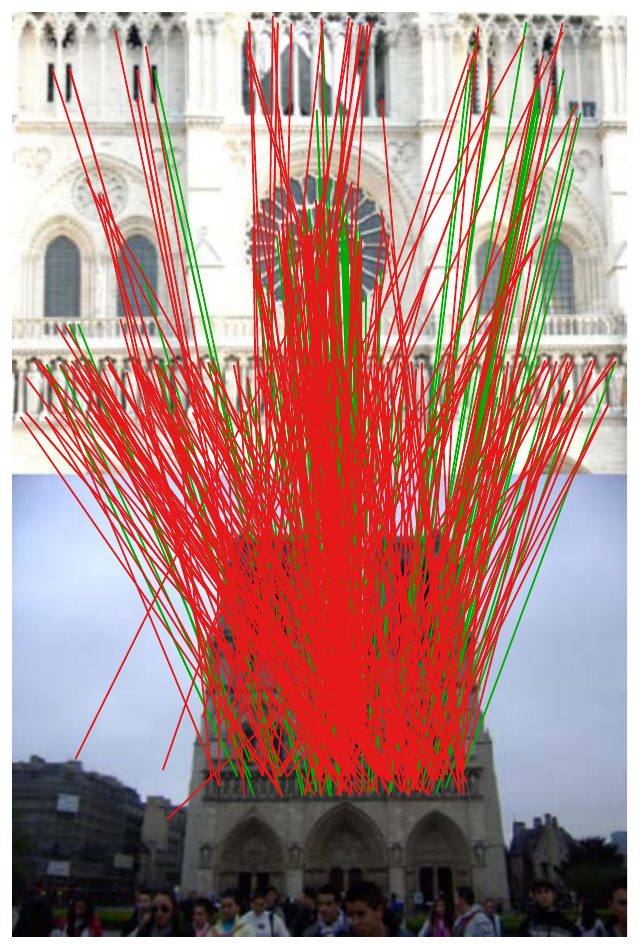}\hspace{0mm} &
  \includegraphics[width=0.495\linewidth, trim=0 10 0 0, clip]{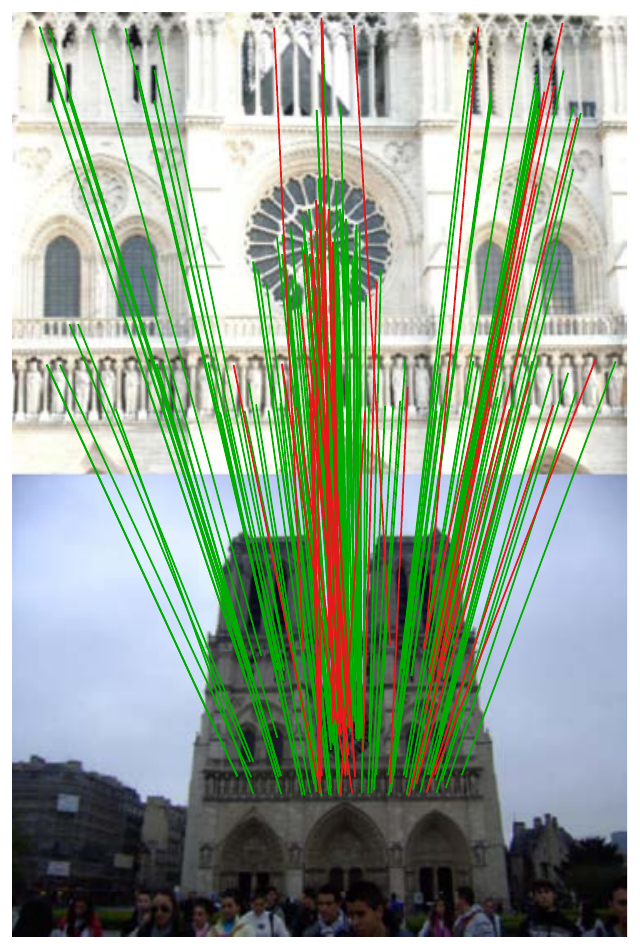}\\
  (a) RANSAC & (b) Our approach \\
\end{tabular}
\vspace{0.5mm}
\caption{We extract $2$k SIFT keypoints from a challenging image pair and
  display the inlier matches found with RANSAC (left) and our approach
  (right). We draw them in green if they conform to the ground-truth epipolar
  geometry, and in red otherwise.}
\label{fig:teaser}
\vspace{-1.2em}
\end{figure}

Recovering the relative camera motion between two images is one of the most
basic tasks in Computer Vision, and a key component of wide-baseline stereo and
Structure from Motion (SfM) pipelines. However, it remains a difficult problem
when dealing with wide baselines, repetitive structures, and illumination
changes, as depicted by \fig{teaser}.
Most algorithms rely on sparse keypoints~\cite{Lowe04,Bay08,Rublee11}
to establish an initial set of correspondences across images, then try to find
a subset of reliable matches---inliers---which conform to a given geometric
model, and use them to recover the pose~\cite{Hartley00}.  They
rely on combinations of well-established techniques, such as
SIFT~\cite{Lowe04}, RANSAC~\cite{Fischler81}, and the 8-point
algorithm~\cite{Longuet-Higgins81}, which have been in use for decades.

With the advent of deep learning, there has been a push towards
reformulating local feature extraction using neural
networks~\cite{Yi16b,Simo-Serra15}. However, while these algorithms outperform
earlier ones on point-matching benchmarks, incorporating them into
pose estimation pipelines may not necessarily translate into a performance
increase, as indicated by two recent studies~\cite{Schonberger17,Bian17}.
This suggests that the limiting factor may not lie in {\it establishing} the
correspondences as much as in {\it choosing} those that are best to recover the pose.

This problem has received comparatively little attention and most algorithms
still rely on non-differentiable hand-crafted techniques to solve
it. DSAC~\cite{Brachmann16b} is the only recent attempt we know of to tackle
sparse outlier rejection in a differentiable manner. However, this method is
designed to mimic RANSAC rather than outperform it.  Furthermore, it is
specific to 3D to 2D correspondences,
rather than stereo.


By contrast, we propose a novel approach to finding geometrically consistent
correspondences with a deep network. 
Given
feature points in both images, and potential correspondences between them, we
train a deep network that simultaneously solves a classification
problem---retain or reject each correspondence---and a
regression problem---retrieving the camera motion---by exploiting
epipolar constraints. To do so we introduce a {\it
  differentiable} way to compute the pose through a simple weighted
reformulation of the 8-point algorithm, with the weights predicting
the likelihood of a correspondence being correct. In practice, we assume the camera
intrinsics to be known, which is often true because they are stored in the
image meta-data, and express 
camera motion in terms of the essential matrix~\cite{Hartley00}.

\begin{figure}
  \centering \includegraphics[width=\linewidth, clip]{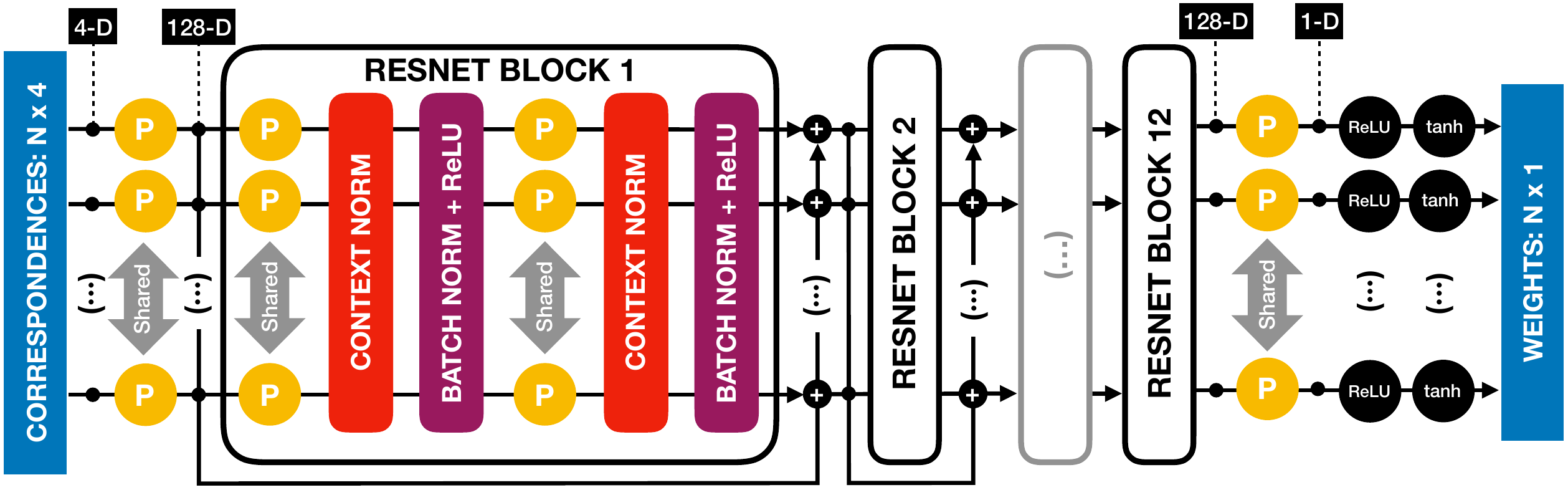}
  \caption{Our deep network takes $N$ correspondences
    between two 2D points~($4 \times N$ values) as input, and produces
    one weight for each correspondence, encoding its
    likelihood to be an inlier.
    Each correspondence is processed independently by weight-sharing
    Perceptrons (P), rendering the network invariant to permutations of the
    input. Global information is embedded by Context Normalization, a novel
    technique we detail in \refsec{network}.}
  \label{fig:architecture}
  \vspace{-1.5em}
\end{figure}

As shown in \fig{architecture}, our architecture relies on Multi Layer
Perceptrons~(MLPs) that are applied independently on each correspondence,
rendering the network invariant to the order of the input. This is inspired by
PointNet~\cite{Qi17}, which runs an MLP on each individual point from a
3D cloud and then feeds the results to an additional network that generates a
global feature, which is then appended to each point-wise feature.
By contrast, 
we simply normalize the
distribution of the feature maps
over all correspondences every time they go through a perceptron.
As the correspondences are constrained by the camera motion, this procedure
provides context.
We call this novel, non-parametric
operation {\it Context Normalization}, and found it simpler and more effective than the
global context features of~\cite{Qi17} for our purposes.

Our method has the following advantages:
(i) it can double the performance of the state of the art;
(ii) being keypoint-based, it generalizes better than image-based dense methods
to unseen scenes, which we demonstrate with a {\it single model} that
outperforms current methods on drastically different indoors and outdoors
datasets;
(iii) it requires only weak supervision through essential matrices for
training;
(iv) it can work effectively with very little training data, \eg, we can still
outperform the state of the art on challenging outdoor scenes with only 59
training images.



\section{Related Work}
\label{sec:related}

\vspace{-1mm}
\paragraph{Traditional handcrafted methods.} The traditional approach for
estimating the relative camera motion between two images is to use sparse
keypoints, such as SIFT~\cite{Lowe04}, to establish an initial set of
correspondences, and reject outliers with RANSAC~\cite{Fischler81}, using the
5-point algorithm~\cite{Nister03} or the 8-point
algorithm~\cite{Longuet-Higgins81} to retrieve the essential matrix.

Many works have focused on improving the outlier rejection step of this
pipeline, \ie, RANSAC. MLESAC~\cite{Torr00} shows improvements when solving
image geometry problems. PROSAC~\cite{Chum05a} speeds up the estimation
process. USAC~\cite{Raguram13} combines multiple advancements together into a
unified framework. Least median of squares~(LMEDS)~\cite{Rousseeuw87} is also
commonly used to replace RANSAC. A comprehensive study on this topic can be
found in~\cite{Choi09, Raguram13}.  In practice, however, RANSAC still remains
to be the {\it de facto} standard.


A major drawback of these approaches is that they rely on small subsets of the
data to generate the hypotheses, \eg, the 5-point algorithm considers only five
correspondences at a time. This is sub-optimal, as image pairs with large
baselines and imaging changes will contain a large percentage of outliers, thus
making most of the hypotheses useless.

Recent works try to overcome this limitation by simultaneously rejecting
outliers and estimating global motion. \cite{Lin14b} assumes that global motion
should be piece-wise smooth.  GMS~\cite{Bian17} further exploits this
assumption by dividing the image into multiple grids and forming initial
matches between the grid cells. Although they show improvements over
traditional matching strategies, the piece-wise smoothness assumption is often
violated in practice.

\vspace{-4mm}
\paragraph{Learning-based methods.} 
Solving image correspondence with deep networks has received a lot
of interest 
recently. In contrast with traditional methods,
many of these new techniques~\cite{Vijayna17,Ummenhofer17,Zhou17,Zamir16} are
dense, using the raw image as input.  They have shown promising results
on constrained stereo problems, such as matching two frames from a video
sequence, but the general problem remains far from solved. As evidenced
by our experiments, this approach can be harmful on scenes with
occlusions or large baselines, which are
commonplace in photo-tourism applications.

Dense methods aside, DSAC~\cite{Brachmann16b} recently introduced a
differentiable outlier rejection scheme for monocular pose estimation.
However, it relies on a strategy to evaluate hypotheses that is particular to
monocular pose estimation and difficult to extend to stereo. Furthermore, this
technique amounts to replacing RANSAC with a differentiable counterpart for
end-to-end training, which does not significantly improve the performance
beyond RANSAC, as we do.

In~\cite{DeTone17}, a dual architecture with separate networks for extracting
sparse keypoints and forming the correspondences, assuming a homography model,
was proposed. Because of its requirement for ground-truth annotations of object
corners, this work was only demonstrated on synthetic, non-realistic images. By
contrast, our method only requires essential matrices for training and thus, as
demonstrated by our experiments, is effective in real-world scenarios.




\section{Method}
\label{sec:method}

Our goal is to establish valid and geometrically consistent correspondences
across images and use them to retrieve the camera motion.  We rely on local
features, which can be unreliable and require outlier rejection. Traditional
methods approach this problem by iteratively sampling small subsets of matches,
which makes it difficult to exploit the global context. By contrast, we develop
a deep network that can leverage all the available information in a single
shot.

Specifically, we extract local features over two images, create an overcomplete
set of putative correspondences, and feed them to the network depicted by
\fig{architecture}. It assigns a weight to each correspondence, which encodes
their likelihood of being an inlier. We further use it to set the influence of
each correspondence in a weighted reformulation of the eight-point
algorithm~\cite{Longuet-Higgins81}. In other words, our method performs joint
inlier/outlier classification and regression to the essential matrix.  The
relative rotation and translation between cameras can then be recovered with a
singular value decomposition of the resulting essential
matrix~\cite{Hartley00}.

In the remainder of this section, we formalize the problem and describe the
architecture of \fig{architecture}. We then reformulate the eight-point
algorithm for weighted correspondences, and discuss our learning strategy with
a hybrid loss.


\subsection{Problem Formulation}
\label{sec:formulation}

Formally, let us consider a pair of images $(\mathbf{I},\mathbf{I}')$.  We
extract local features $(\mathbf{k}_i,\mathbf{f}_i)_{i\in[1,N]}$ from image
$\mathbf{I}$, where $\mathbf{k}_i=(u_i,v_i)$ contains keypoint locations, and
$\mathbf{f}_i$ is a vector encoding a descriptor for the image patch centered
at $(u_i,v_i)$. The keypoint orientation and scale are used to extract these
descriptors and discarded afterwards.  Similarly, we obtain
$(\mathbf{k}'_j,\mathbf{f}'_j)_{j\in[1,N']}$ from $\mathbf{I'}$, keeping $N=N'$
for simplicity. Our formulation is thus amenable to traditional features such
as SIFT~\cite{Lowe04} or newer ones such as LIFT~\cite{Yi16b}.

We then generate a list of $N$ putative correspondences by matching each keypoint
in $\mathbf{I}$ to the nearest one in $\mathbf{I'}$ based on the descriptor
distances. We denote this list as
\setlength{\abovedisplayskip}{4pt}
\setlength{\belowdisplayskip}{4pt}
\begin{eqnarray}
\bx & = &    [\bcorr_1, ..., \bcorr_N] \; , \label{eq:netInput} \\
\bcorr _i & = & [u_i,v_i,u'_i,v'_i] \; . \nonumber
\end{eqnarray}
More complex
strategies~\cite{Lowe04,Cho14} could be used, but we found this one
sufficient.
As many traditional algorithms, we use the camera intrinsics to normalize the
coordinates to $[-1,1]$, which makes the optimization numerically
better-behaved~\cite{Hartley00}.
We discard the descriptors, so that
the list of $N$ location quadruplets is the only input to our network.

As will be discussed in \refsec{inference}, it is straightforward to extend the
eight-point algorithm to take as input a set of correspondences with
accompanying weights
\begin{equation}
\bw =  [w_1, ..., w_N] \; ,
\label{eq:netOutput}
\end{equation}
where $w_i \in [0,1]$ is the score assigned to correspondence $\bcorr_i$,
with $w_i=0$ standing for $\bcorr_i$ being an outlier.
Then, let $g$ be a function that takes as input $\bx$ and $\bw$ and returns the
essential matrix $\bE$.  
Now, given a set of $P$ image pairs
$(\mathbf{I}_k,\mathbf{I}_k')_{k\in[1,P]}$
and their corresponding essential matrices $\bE_k$,
we extract the set of correspondences $\bx_k$ of
Eq.~\eqref{eq:netInput}, and our problem becomes that of designing a deep
network that encodes a mapping $f$ with parameters $\Phi$,
such that
\begin{eqnarray}
 \forall \; k \; ,  1 \leq k \leq P \; , \; \bw_{k}  =  f_\Phi\left( \bx_k\right) \;, \label{eq:netMapping} \\
  \bE_k \approx g(\bx_k,\bw_k) \;.  \nonumber
\end{eqnarray}



\subsection{Network Architecture}
\label{sec:network}

We now describe the network that implements the mapping
of Eq.~\eqref{eq:netMapping}. Since the order of the correspondences is
arbitrary, permuting $\bx_k$ should result in an equivalent permutation of
$\bw_k=f_{\Phi}(\bx_k)$. To achieve this, inspired by PointNet~\cite{Qi17}, a
deep network designed to operate on unordered 3D point clouds, we 
exploit Multi-Layer Perceptrons (MLP). As MLPs operate on {\it individual}
correspondences, unlike convolutional or dense networks, incorporating
information from other perceptrons, \ie, the context, is {\it indispensable}
for it to work properly.
The distinguishing factor between PointNet and our method is how this is done.

In PointNet, the point-wise features are explicitly
combined by a separate network that generates a global context feature, which
is then concatenated to each point-wise output.
By contrast, as shown in \fig{architecture}, we introduce a simple strategy to
normalize the feature maps according to their distribution, after every
perceptron.  This lets us process each correspondence
separately while framing it in the global context defined by the rest,
which encodes camera motion and scene geometry.  We call this non-parametric
operation {\it Context Normalization} (CN).

Formally, let $\bo_i^{l} \in \IR^{C^l}$ be the output of layer $l$ for
correspondence $i$, where $C^l$ is the number of neurons in $l$. We take the
normalized version of $\bo_i^{l}$ to be
\begin{equation}
  \text{CN}\left(\bo_i^{l}\right) =
  \frac{(\bo_i^{l} - \bmu^{l})}{\bsigma^{l}}
  \;\;,
\end{equation}
where
\begin{equation}
  \bmu^{l} = \frac{1}{N}\sum_{i=1}^{N}\bo_i^{l}
  \;,\;\;
  \bsigma^{l} =
  \sqrt{\frac{1}{N}\sum_{i=1}^{N}\left(\bo_i^{l}-\bmu^{l}\right)^2}
  \;\;.
\end{equation}

This operation is mechanically similar to other normalization
techniques~\cite{Ioffe15,Ba16,Ulyanov16}, but is applied to a different
dimension and plays a different role. 
We normalize each perceptron's output across
correspondences, but separately for each image pair.
This allows the distribution of the feature maps to
encode scene geometry and camera motion,
embedding contextual information into context-agnostic MLPs.

By contrast, the other normalization techniques primarily focus on
  convergence speed,
with little impact on how the networks operate. Batch
Normalization~\cite{Ioffe15} normalizes the input to each neuron over a
mini-batch, so that it follows a consistent distribution while training. Layer
Normalization~\cite{Ba16} transposes this operation to channels, thus being
independent on the number of observations for each neuron.
They do {\it not}, however, add
contextual information to the input. Batch Normalization assumes that every
sample follows the same global statistics, and reduces to subtracting and
dividing by fixed scalar values at test time. Layer Normalization is applied
independently for each neuron, \ie, it considers the features for one
correspondence at a time.

More closely related is Instance Normalization~\cite{Ulyanov16}, which applies
the same operation we do to full images to normalize their contrast, for image
stylization. However, it is also limited to enhancing convergence, as in their
application context is already captured by spatial convolutions, which are not
amenable to the sparse data in our problem.

Note however that our technique is compatible with Batch Normalization, and we
do in fact use it to speed up training.  As shown in \fig{architecture}, our
network is a 12-layer ResNet~\cite{He16}, where each layer contains two
sequential blocks consisting of: a Perceptron with 128 neurons sharing weights
for each correspondence, a Context Normalization layer, a Batch Normalization
layer, and a ReLU.  After the last perceptron, we apply a ReLU followed by a
$\tanh$ to force the output in the range $\left[0, 1\right)$.  We use this
truncated $\tanh$ instead of, \eg, a sigmoid, so that the network can easily
output $w_i=0$ to completely remove an outlier.



\subsection{Computing the Essential Matrix}
\label{sec:inference}

We now define the function $g$ of Eq.~\eqref{eq:netMapping} that estimates the
essential matrix from a weighted set of correspondences.  If we are to
integrate it into a Deep Learning framework, it must be differentiable with
respect to the weights. We first outline the 8-point
algorithm~\cite{Longuet-Higgins81} that tackles the unweighted scenario, and
then extend it to the weighted case.

\vspace{-3mm}
\paragraph{Traditional Formulation.}
Let $\bx$ be a set of $N$ correspondences $\bcorr_i = [u_i,v_i,u_i',v_i']$.
When $N \geq 8$, the essential matrix $\bE \in \mathbb{R}^{3\times 3}$ can be
computed by solving a singular value problem~\cite{Hartley00} as follows. We
construct a matrix $\bX \in \IR^{N\times9}$ each row of which is computed from a
different correspondence and has the form
$\left[uu', uv', u, vu', vv', v, u', v', 1 \right]$, and we reorganize the
coefficients of $\bE$ into a column vector $\vectorize\left(\bE\right)$. This
vector has to be the unit vector that minimizes
$\left\|\bX\vectorize\left(\bE\right)\right\|$, or equivalently
$\left\|\bX^\top\bX\vectorize\left(\bE\right)\right\|$. Therefore,
$\vectorize\left(\bE\right)$ is the eigenvector associated to the smallest
eigenvalue of $\bX^\top\bX$. Additionally, as the essential matrix needs to
have rank 2, we find the rank-2 matrix $\hat{\bE}$ that minimizes the Frobenius
norm $\|\bE - \hat{\bE}\|_F$~\cite{Hartley00}.

\vspace{-3mm}
\paragraph{Weighted reformulation.}
In practice, the 8-point algorithm can be severely affected by outliers and is
used in conjunction with an outlier rejection scheme.  In our case, given the
weights $\bw$ produced by the network, we can simply minimize
$\left\|\bX^\top\diag\left(\bw\right)\bX\vectorize\left(\bE\right)\right\|$
instead of $\left\|\bX^\top\bX\vectorize\left(\bE\right)\right\|$, where
$\diag$ is the diagonalization operator. This amounts to giving weight $w_i$ to
the $i$-th row in $\bX$, representing the contribution of the $i$-th
correspondence.

Since $\bX^\top\diag\left(\bw\right)\bX$ is symmetric, the function $g$ that
computes $\vectorize\left(\bE\right)$, and therefore $\bE$, from $\bX$ is a
self-adjoint eigendecomposition, which has differentiable closed-form solution
with respect to $\bw$~\cite{Ionescu15}. Note that $g(\bx,\bw)$ only depends on
the inliers because any correspondence with zero weight has strictly no
influence on the result.  To compute the final essential matrix $\hat{\bE}$, we
follow the same procedure as in the unweighted case, which is also
differentiable.


\newcommand{\imh}{1.9cm}
\newcommand{\imsp}{1mm}
\begin{figure*}
\centering
\small
\begin{tabular}{@{}c@{}c@{}c@{}c@{}c@{}c@{}c@{}}
\includegraphics[height=\imh]{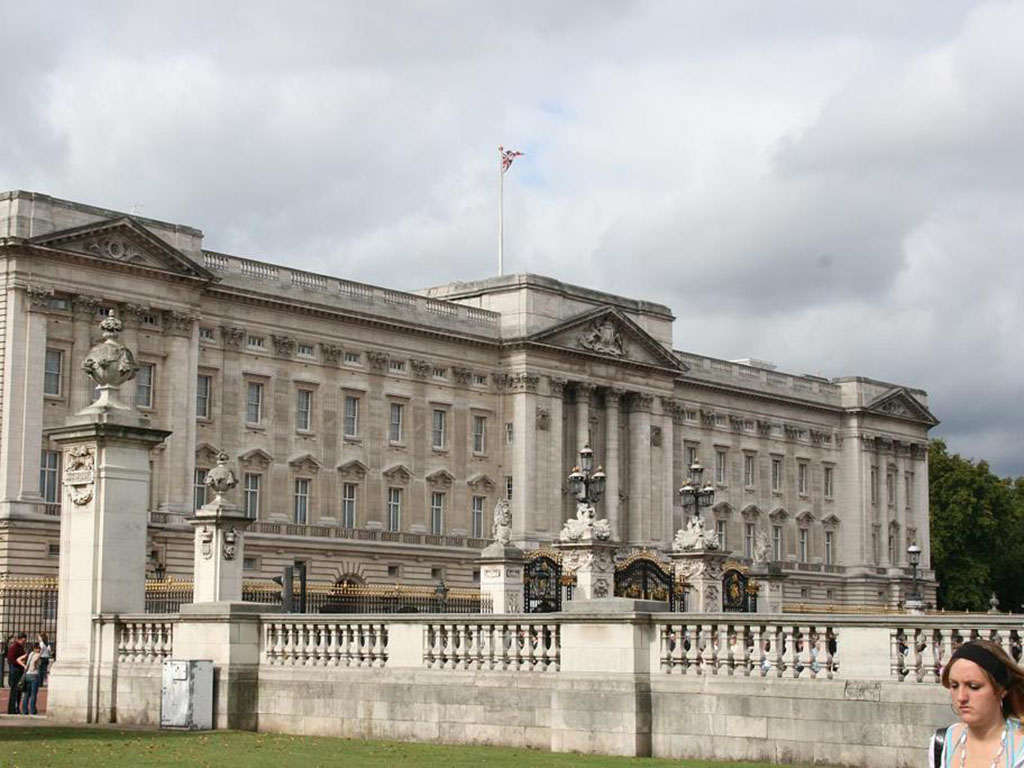}\hspace{\imsp} &
\includegraphics[height=\imh]{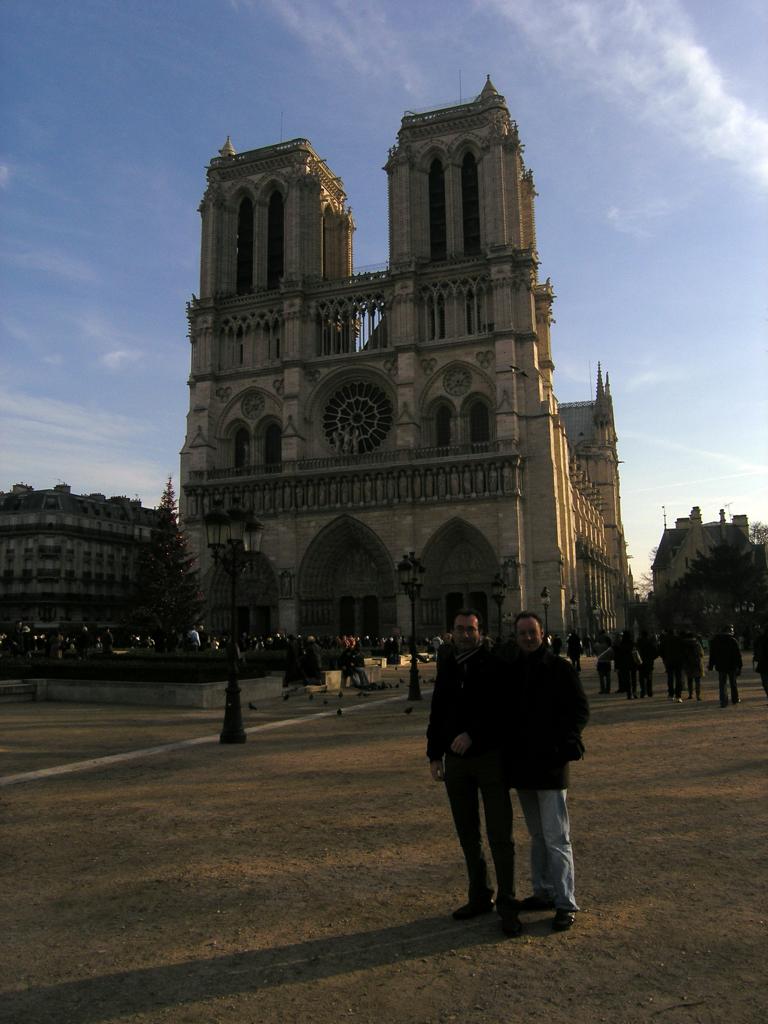}\hspace{\imsp} &
\includegraphics[height=\imh]{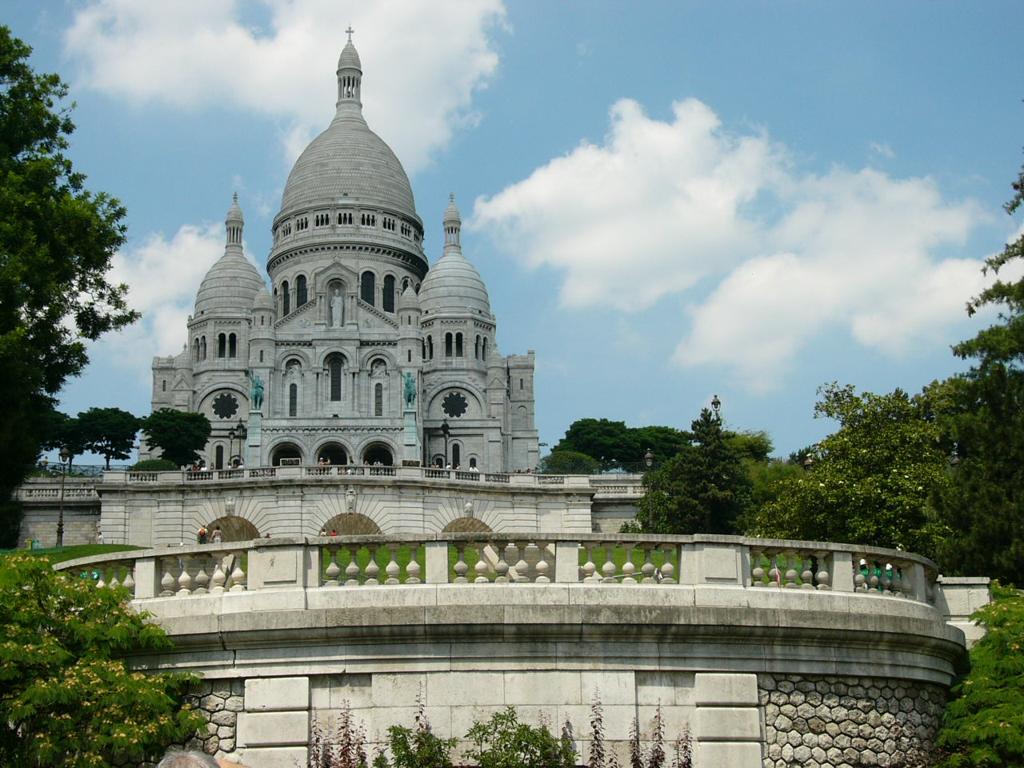}\hspace{\imsp} &
\includegraphics[height=\imh]{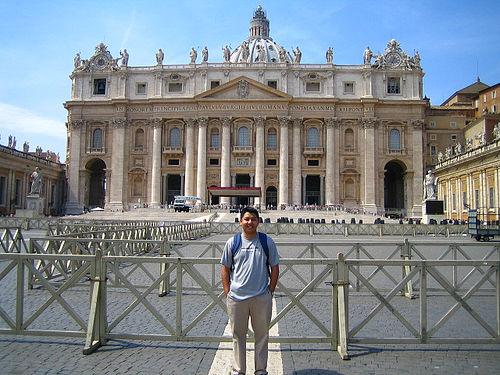}\hspace{\imsp} &
\includegraphics[height=\imh]{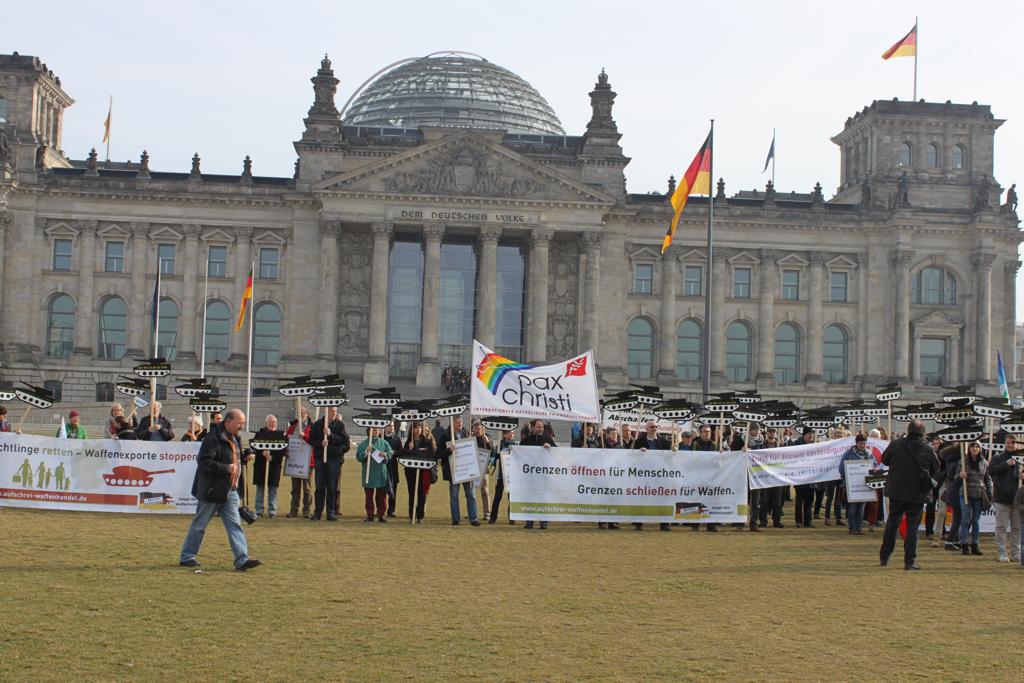}\hspace{\imsp} &
\includegraphics[height=\imh]{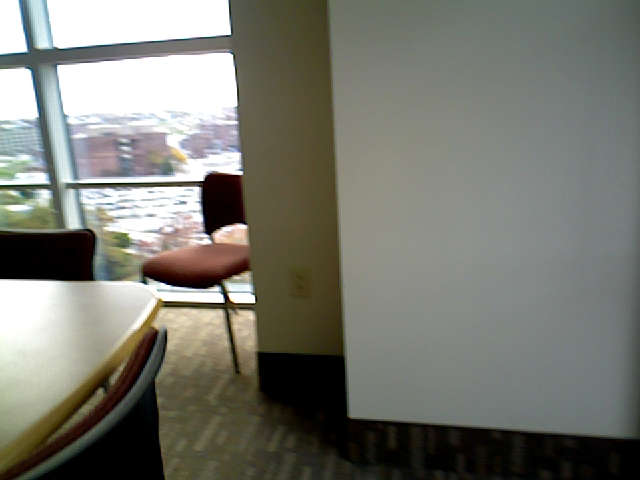}\hspace{\imsp} &
\includegraphics[height=\imh]{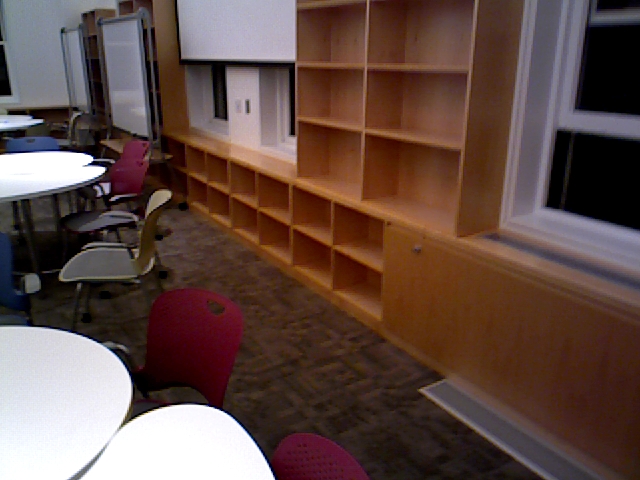} \\
\includegraphics[height=\imh]{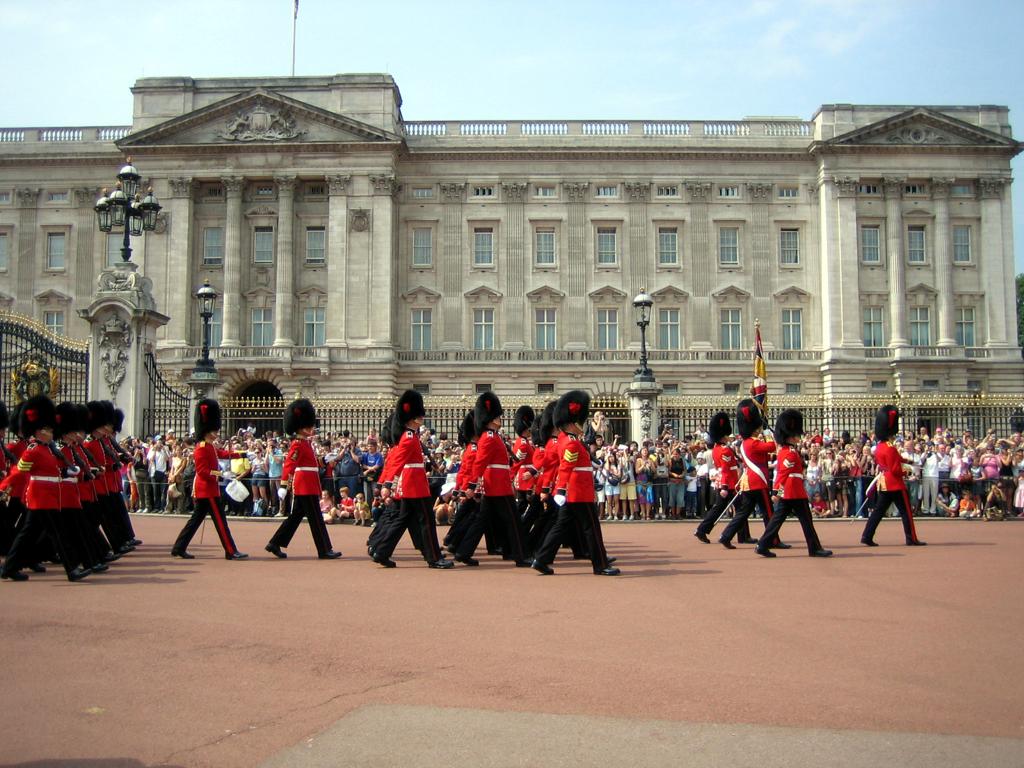}\hspace{\imsp} &
\includegraphics[height=\imh]{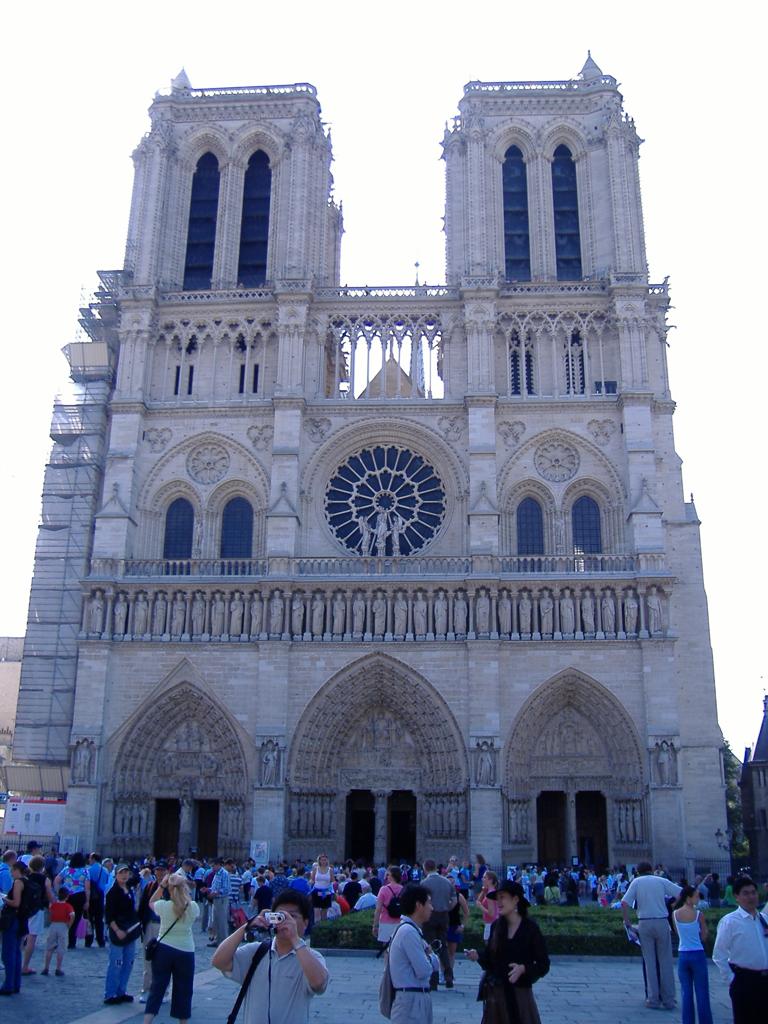}\hspace{\imsp} &
\includegraphics[height=\imh]{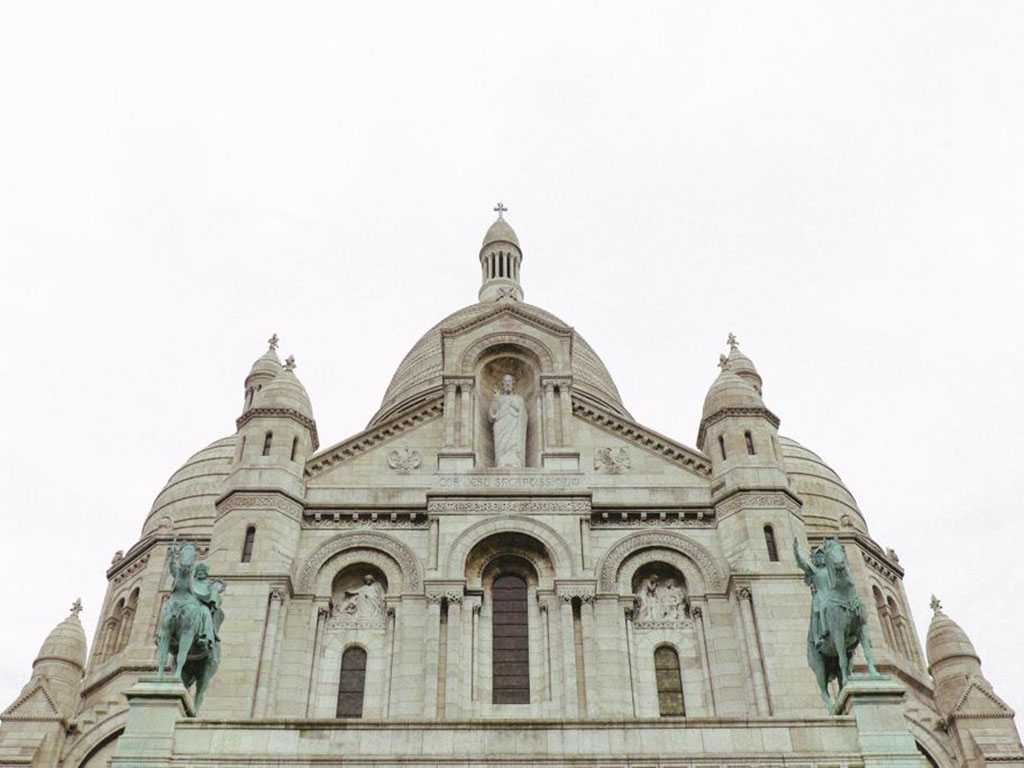}\hspace{\imsp} &
\includegraphics[height=\imh]{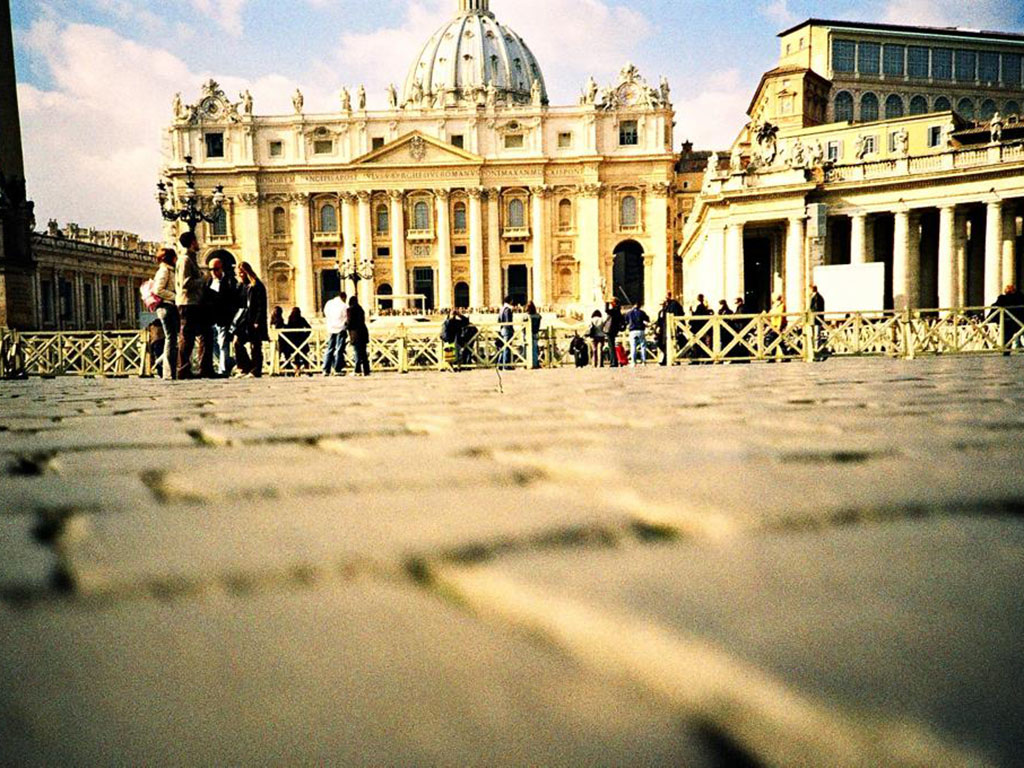}\hspace{\imsp} &
\includegraphics[height=\imh]{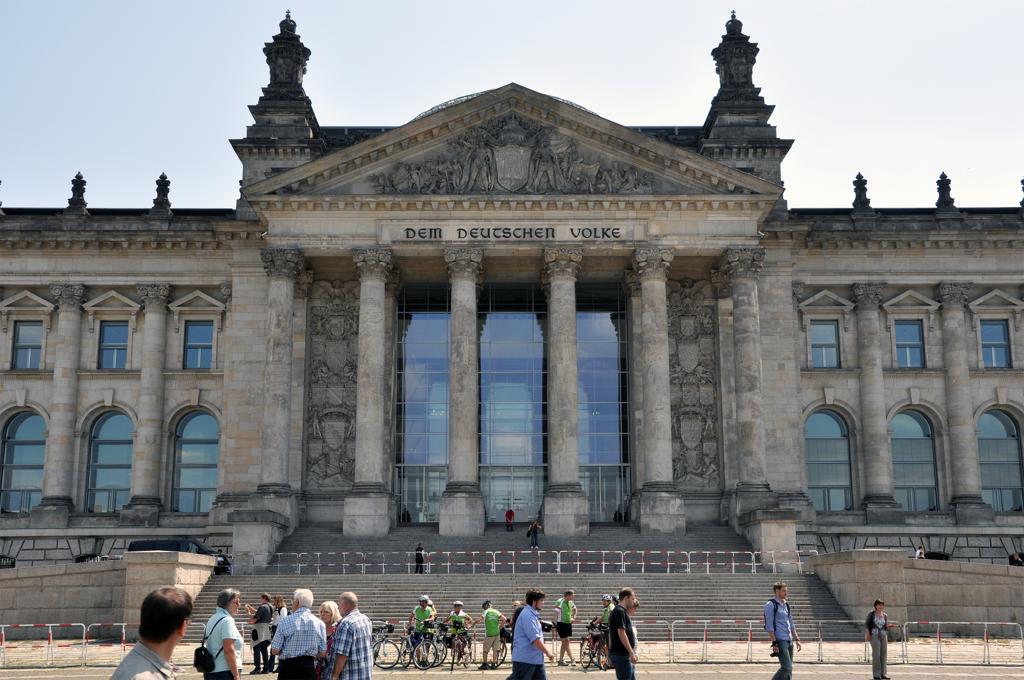}\hspace{\imsp} &
\includegraphics[height=\imh]{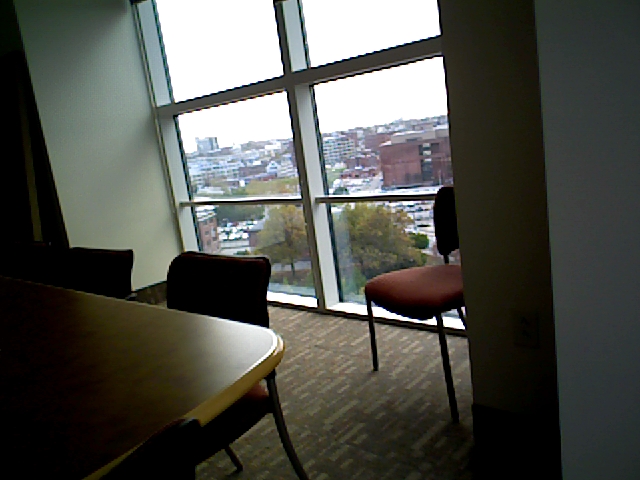}\hspace{\imsp} &
\includegraphics[height=\imh]{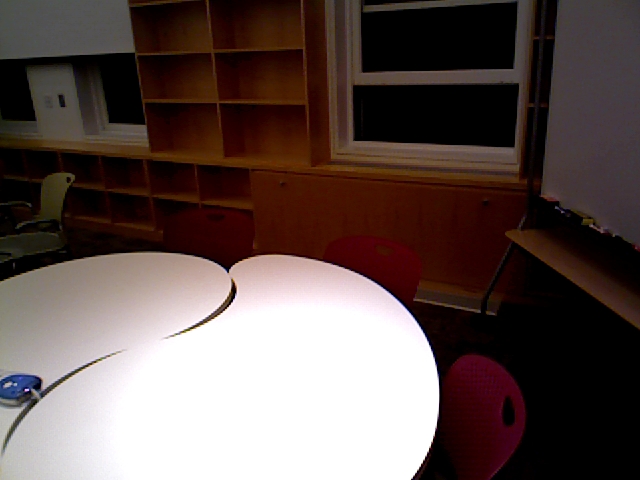} \\
\end{tabular}
\caption{Matching pairs from our training sets. Left to right:
`Buckingham', `Notredame', `Sacre Coeur', `St. Peter's' and `Reichstag', from~\cite{Thomee16,Heinly15};
and `Brown 1' and `Harvard 1' from \cite{Xiao13} (please see the appendix for details). Note that we crop the images for
presentation purposes, but our algorithm is based on sparse correspondences and is
thus not restricted in any way by image size or aspect ratio.}
\label{fig:pair-examples}
\vspace{-3mm}
\end{figure*}


\subsection{Learning with Classification and Regression}
\label{sec:optim}

\noindent%
We train the network $f_{\Phi}$ with a hybrid loss function
\begin{equation}
  \loss(\Phi) = \sum_{k=1}^P (\alpha \loss_{x}(\Phi,\bx_k) + \beta \loss_{e}(\Phi,\bx_k)) \; ,
\label{eq:hybrid_loss}
\end{equation}
where $\Phi$ are the network parameters and $\bx_k$ is the set of putative
correspondences for image pair $k$. $\loss_{x}$ is a classification loss
aggregated over each individual correspondence, and $\loss_{e}$ is a regression
loss over the essential matrix, obtained from the weighted 8-point algorithm of
\refsec{inference} with the weights produced by the network. $\alpha$ and
$\beta$ are the hyper-parameters weighing the two loss terms.

\vspace{-3mm}
\paragraph{Classification loss $\loss_{x}$: rejecting outliers.}

Given a set of $N$ putative correspondences $\bx_k$ and their respective labels
$\by_k = \left[y_k^1, ..., y_k^N \right]$, where $y_k^i \in \{0,1\}$ and
$y_k^i=1$ denotes that the $i$th correspondence is an inlier, we define
\begin{equation}
  \loss_{x}(\Phi,\bx_k)= \frac{1}{N}\sum_{i=1}^{N}\gamma_{k}^iH\left(y_k^i, S\left(o_k^i\right)\right) \; ,
  \label{eq:classification_loss}
\end{equation}
where $o_k^i$ is the linear output of the last layer for the $i$-th
correspondence in training pair $k$,
\kyi{
$S$ is the logistic function used in conjunction with the binary cross entropy
$H$,
}
and $\gamma_{k}^i$ is the per-label weight to balance positive and
negative examples.

To avoid exhaustive annotation, we generate the labels
$\by$ by exploiting the epipolar constraint.
Given a point in one image, if the corresponding point in the other image does
not lie on the epipolar line, the correspondence is spurious.
We can quantify this with the epipolar distance~\cite{Hartley00}
\begin{equation}
  d(\bp,\bE\bp')= \frac{\bp'^T \bE \bp}{\sqrt{(\bE \bp)^2_{[1]} + (\bE \bp)^2_{[2]}}} \; ,
\label{eq:epipolar_distance}
\end{equation}
%
where $\bp=[u,v,1]^T$ and $\bp'=[u',v',1]^T$ indicate a putative correspondence
between images $\bI$ and $\bI'$
in homogenous coordinates, and $\mathbf{v}_{[j]}$ denotes the $j$th element of
vector $\mathbf{v}$.  In practice, we use the {\it symmetric epipolar distance}
$d(\bp,\bE\bp') + d(\bp',\bE^T\bp)$,
with a threshold of $10^{-2}$ in normalized coordinates to determine valid correspondences.

Note that this is a weak supervisory signal because outliers can still have a
small epipolar distance if they lie close to the epipolar line. However, a
fully-supervised labeling would require annotating correspondences for every point in both images, which is
prohibitively expensive, 
or depth data,
which is not available in most scenarios. Our experiments show that, in
practice, weak supervision is sufficient.

\vspace{-3mm}
\paragraph{Regression loss $\loss_{e}$: predicting the essential matrix.}
The weak supervision applied by the classification loss proves robust in our
experience, but it may still let outliers in. We propose to improve it by using
the weighted set of correspondences to extract the essential matrix and
penalize deviations from the ground-truth. We therefore write
\begin{equation}
  \loss_{e}(\Phi,\bx_k) = \min \left\{
    \left\|\bE_k^* \pm g\left(\bx_k,\bw_k\right)\right\|^2
  \right\}
\;,
\label{eq:essential_loss}
\end{equation}
where $\bE_k^*$ is the ground-truth essential matrix, and $g$ is the function
defined in Section~\ref{sec:inference} that predicts an essential matrix from a
set of weighted correspondences. We have to compute either the difference or
the sum because the sign of $g(\bx,\bw)$ can be flipped with respect to that of
$\bE^*$. Note that here we use the non-rank-constrained estimate of the
essential matrix $\bE$, instead of $\hat{\bE}$, as defined in
\refsec{inference}, because minimizing $\loss_{e}$ already aims to get the
correct rank.

\vspace{-3mm}
\paragraph{Optimization.}
We use Adam~\cite{Kingma15} to minimize the loss $\loss(\Phi)$, with a learning
rate of $10^{-4}$ and mini-batches of 32 image pairs. The classification loss
can train accurate models by itself, and we observed that using the regression
loss early on can actually harm performance. We find empirically that setting
$\alpha=1$ and enabling the regression loss after $20$k batches with
$\beta=0.1$ works well. This allows us to increase relative performance by 5 to
20\%.



\subsection{RANSAC at Test Time}
\label{sec:ransac}

The loss function must be differentiable with respect to the
network weights. However, this requirement disappears at test time. We can
thus apply RANSAC on the correspondences labeled as inliers by our
network
to weed out any remaining
outliers.  We will show that this performs
much better stand-alone RANSAC, which
confirms our intuition that sampling is a sub-optimal way to approach
``needle-in-the-haystack'' scenarios with a large ratio of outliers.




\section{Experimental Results}

We first present the datasets and evaluation protocols, and then justify our
implementation choices. Finally, we benchmark our approach against
state-of-the-art techniques.

\subsection{Datasets}
\label{sec:datasets}

Sparse feature matching is particularly effective in outdoor scenes, and we will
see that this is where our approach shines. By contrast, keypoint-based methods
are less suitable for indoor scenes. Nevertheless, recent techniques have
tackled this problem, and we show that our approach still outperforms the state
of the art in this scenario.

\vspace{-3mm}
\paragraph{Outdoor scenes.}
To evaluate our method in outdoor settings, we aim to leverage many images
featuring similar scenes seen from different viewpoints. As such, photo-tourism
datasets are natural candidates. We rely on Yahoo's YFCC100M
dataset~\cite{Thomee16}, a collection of 100 million publicly accessible Flickr
images with accompanying metadata, which were later curated~\cite{Heinly15}
into 72 image collections suitable for Structure from Motion (SfM).
We pick five sequences,
depicted in \fig{pair-examples}.  We use
VisualSFM~\cite{Wu13} to recover the camera poses and generate the
ground-truth.

We also consider the `Fountain' and `HerzJesu' sequences
from~\cite{Strecha08b}. They are very small and we only use them for testing,
to show that our networks trained on photo-tourism datasets can successfully
generalize.

\vspace{-4mm}
\paragraph{Indoor scenes.} 
We use the SUN3D dataset~\cite{Xiao13}, which comprises a series of indoor
videos captured with a Kinect, with 3D reconstructions. They feature
office-like scenes with few distinctive features, many repetitive elements, and
substantial self-occlusions, which makes them extremely challenging for sparse
keypoint methods. We select 9 sequences for training and testing, and use 15
sequences previously chosen by~\cite{Ummenhofer17} for testing only, to provide
a fair comparison. We subsample the videos by a factor of 10.

For every sequence we train on, we randomly split the images into disjoint
subsets for training (60\%), validation (20\%) and test (20\%).  To select
valid image pairs on the `Outdoors' subset, we sample two images randomly and
check if they have a minimum number of 3D points in common from the SfM
reconstruction, indicating that the problem is solvable. For the `Indoors' set,
we use the depth maps to determine visibility. We use $1$k image pairs for
validation and testing, and as many as possible for training.

\begin{figure}[!t]
\centering
\includegraphics[width=.9\linewidth]{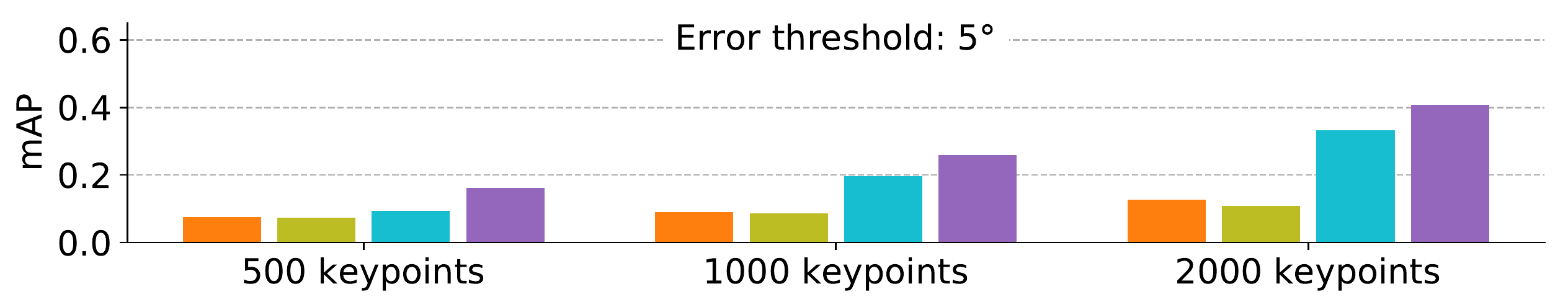}\\
\includegraphics[width=.9\linewidth]{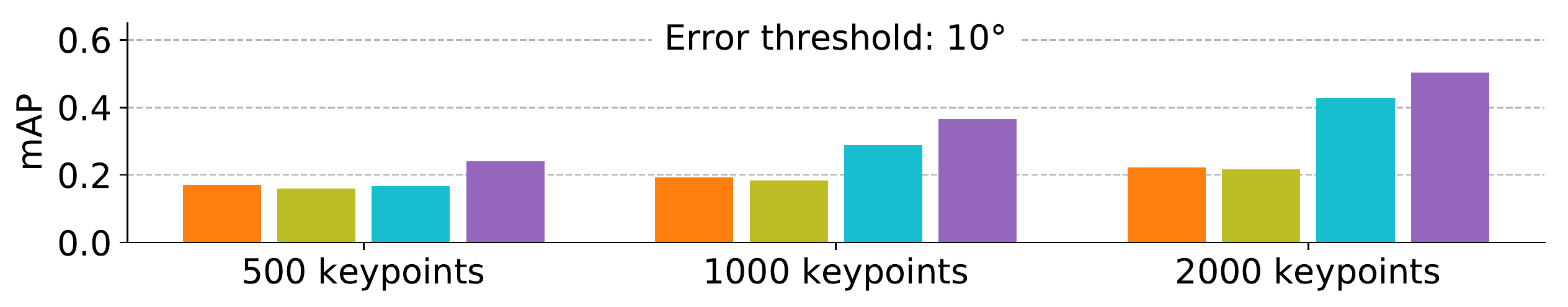}\\
\includegraphics[width=.9\linewidth]{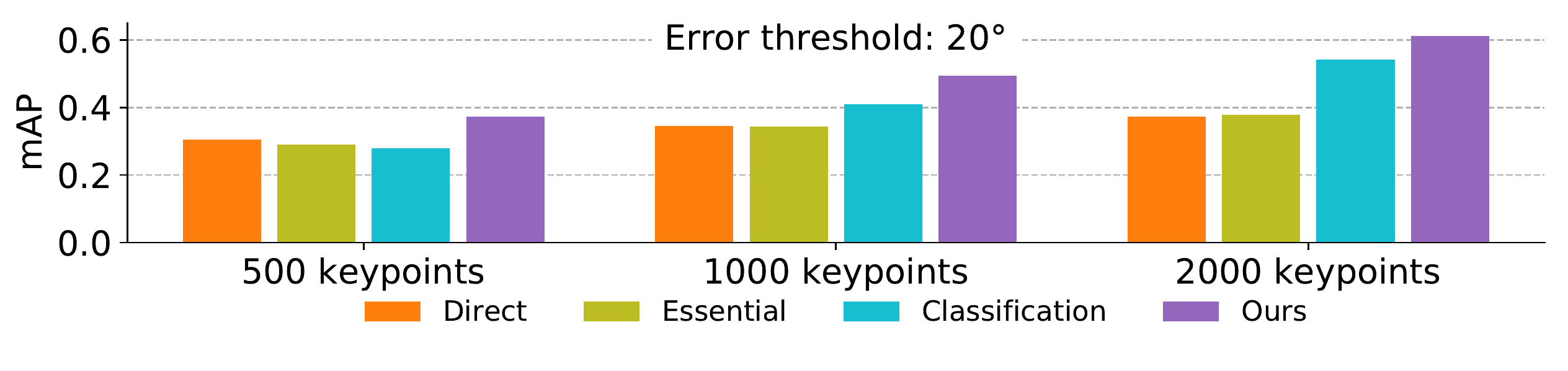}
\caption{mAP for multiple error thresholds and number of keypoints, using the
  four optimization strategies of Section~\ref{sec:variants}.}
\label{fig:results_loss}
\vspace{-3mm}
\end{figure}

\subsection{Evaluation Protocols}

\vspace{0mm}
\paragraph{Keypoint-based methods.} They include well-established algorithms,
RANSAC~\cite{Fischler81}, MLESAC~\cite{Torr00}, and LMEDS~\cite{Rousseeuw87},
as well as the very recent GMS~\cite{Bian17}.  For GMS, we incorporate an
additional RANSAC step as described for our method in \secref{ransac}, which we
empirically found mandatory to obtain good performance.

Note that GMS operates with a large ($10$k) pool of ORB
features~\cite{Rublee11}, and thus behaves similarly to dense methods. For all
others, ours included, we evaluate both SIFT~\cite{Lowe04} and
LIFT~\cite{Yi16b} features. For SIFT we use the OpenCV library, and for LIFT
the publicly available models, which were trained on photo-tourism data
different from ours.

\vspace{-4mm}
\paragraph{Dense methods.} We consider G3DR~\cite{Zamir16} and
DeMoN~\cite{Ummenhofer17}. For G3DR, we implement their architecture and train
it using only the pose component of their loss function, as they argue that
pose estimation is more accurate without the classification
loss~\cite{Zamir16}. For DeMoN, we use the publicly available models, which
were trained on 
SUN3D sequences and on
SfM reconstructions of outdoors sets.

\vspace{-3mm}
\paragraph{Metrics.} Given two images, it is possible to estimate rotation
exactly---in theory---and translation only up to a scale
factor~\cite{Hartley00}. We thus use the angular difference between the
estimated and ground-truth vectors, \ie, the closest arc distance, in
degrees, for both, as our error metric.
We do so as follows. First, we generate a curve by classifying
each pose as accurate or not, \ie we compute the precision, given a
threshold (0 to 180$^o$), and build a normalized cumulative curve as in
\cite{Crivellaro17,Bian17}. Second, we compute the area under this curve (AUC)
up to a maximum threshold of 5, 10 or 20$^o$, because after a point it does not
matter how inaccurate pose estimates are. As the curve measures precision, its
AUC is equivalent to mean average precision (mAP). We apply the same threshold
over rotation and translation, for simplicity.

\begin{figure}[!t]
\centering
\includegraphics[width=.9\linewidth]{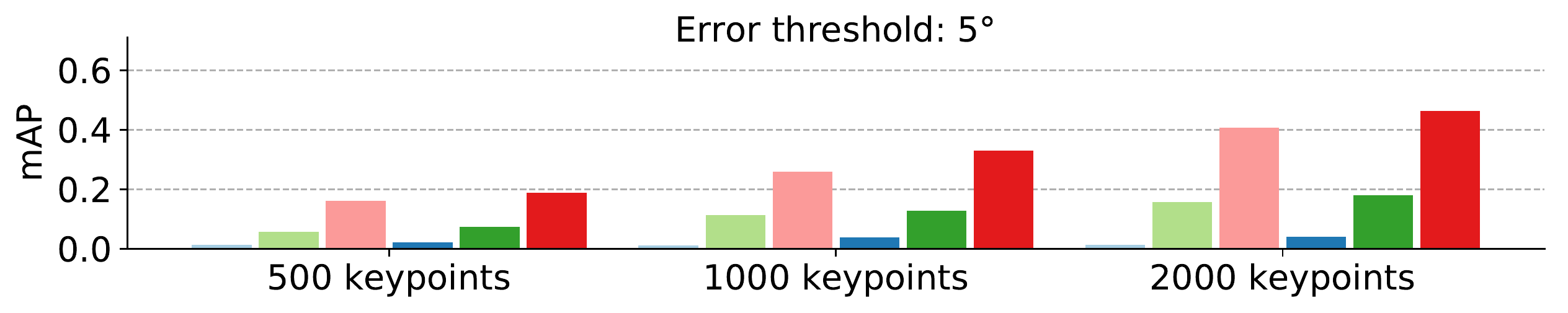}\\
\includegraphics[width=.9\linewidth]{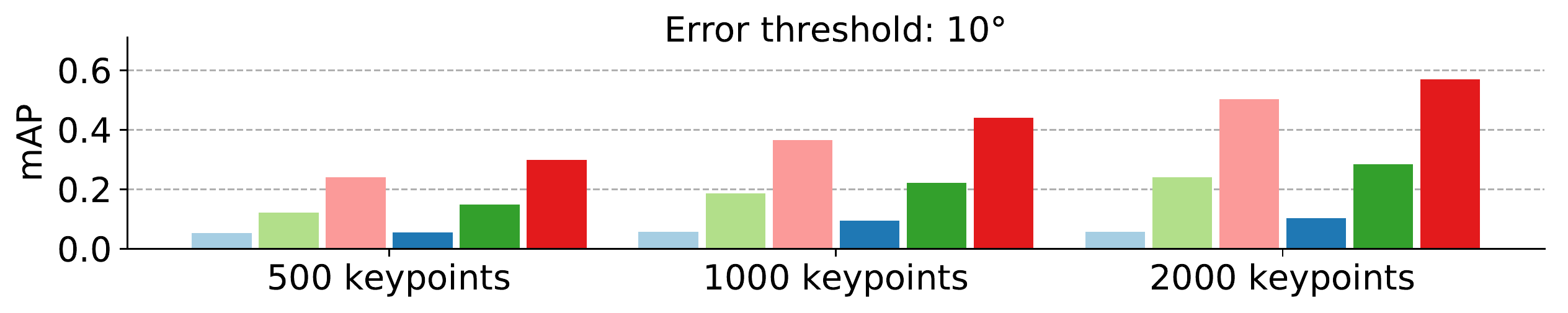}\\
\includegraphics[width=.9\linewidth]{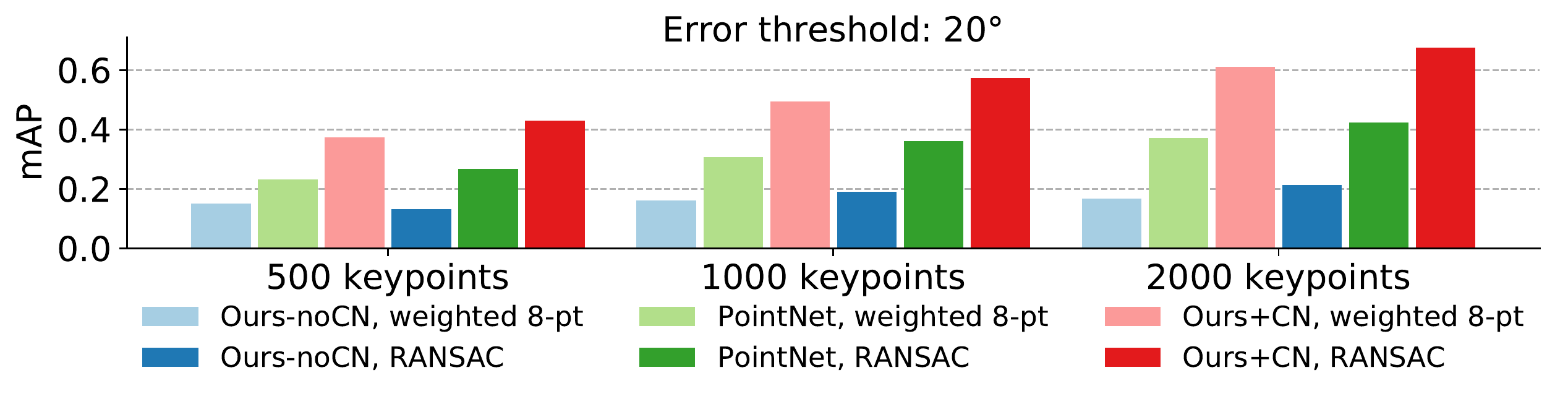}%
\caption{mAP for multiple error thresholds and number of keypoints,
comparing PointNet with our approach, with and without Context Normalization (CN), as explained in \secref{results_architecture}.}
\label{fig:results_architecture}
\vspace{-3mm}
\end{figure}


\begin{figure*}
\centering
\begin{tabular}{@{}cc@{}}
  \includegraphics[width=0.49\linewidth]{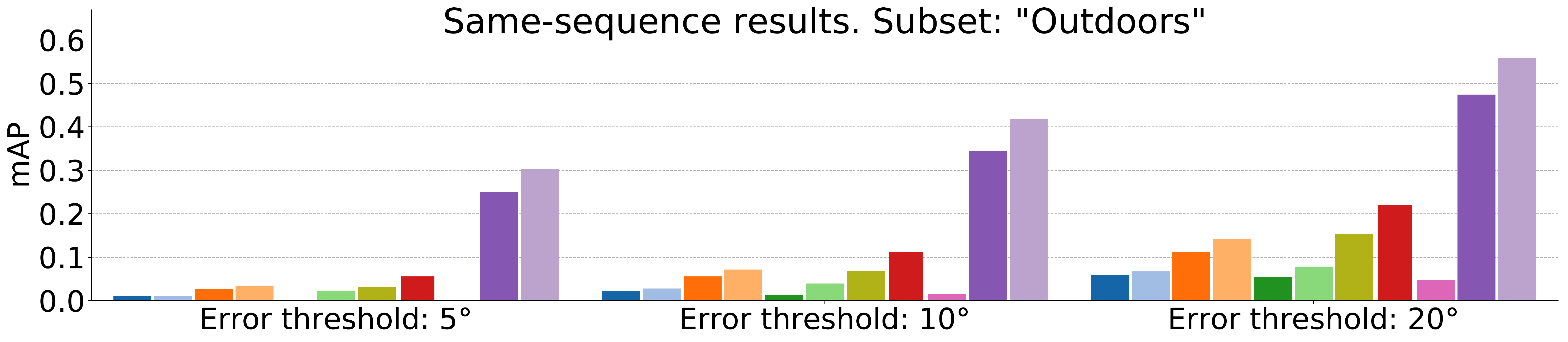} &
  \includegraphics[width=0.49\linewidth]{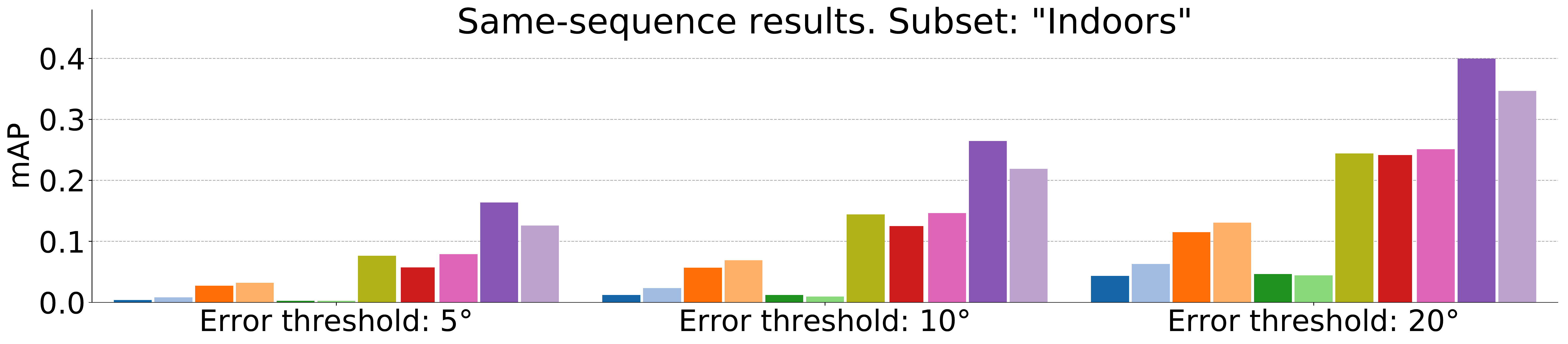} \\
  \multicolumn{2}{c}{\includegraphics[width=0.8\linewidth]{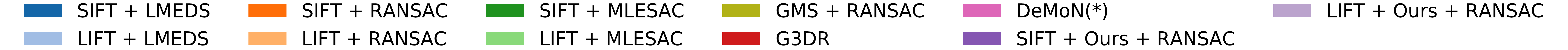}}
\end{tabular}
\caption{Results on known scenes: `Outdoors' (left) and `Indoors' (right). We
  split the images for each set with 60\% for training, 20\% for validation and
  20\% for testing, and report results over 1000 image pairs (or every possible
  combination for smaller sets) for the test split. We mark DeMoN with an
  asterisk as we use the pre-trained models provided by the authors.}
\label{fig:results_sameset}
\end{figure*}


\begin{figure*}
\centering
\begin{tabular}{@{}cc@{}}
  \includegraphics[width=0.49\linewidth]{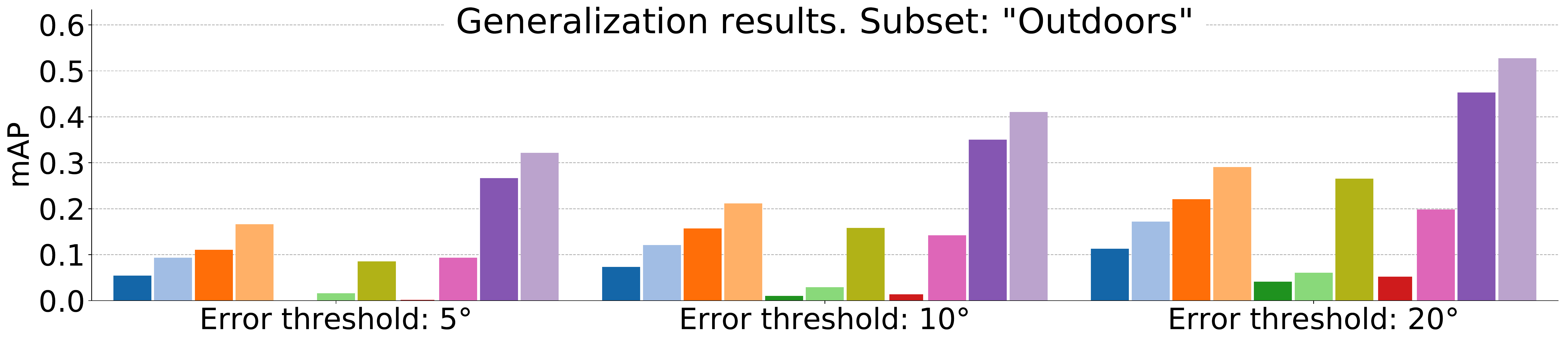} &
  \includegraphics[width=0.49\linewidth]{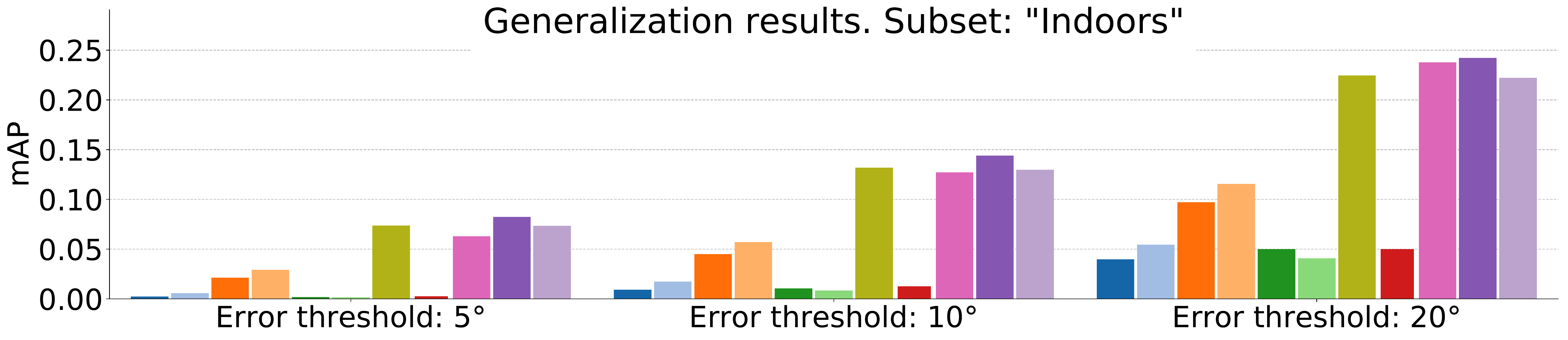} \\
  \multicolumn{2}{c}{\includegraphics[width=0.8\linewidth]{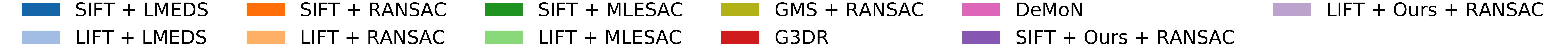}}
\end{tabular}
\caption{Generalization results with models trained and tested on different scenes: `Outdoors' (left) and `Indoors' (right). We train {\em a single model} combining one sequence from the `Indoors' set and one from the `Outdoors' set. For `Outdoors', we test on every other available sequence and average the results. For `Indoors', we test on the 15 sequences chosen by~\cite{Ummenhofer17} for this purpose and average the results.}
\label{fig:results_generalization}
\vspace{-1em}
\end{figure*}

\renewcommand{\imh}{3.02cm}
\renewcommand{\imsp}{0mm}
\begin{figure*}
\centering
\small
\renewcommand{\arraystretch}{.1}
\begin{tabular}{@{}c@{}c@{}c@{}c@{}c@{}c@{}c@{}c@{}c@{}}
\includegraphics[height=\imh]{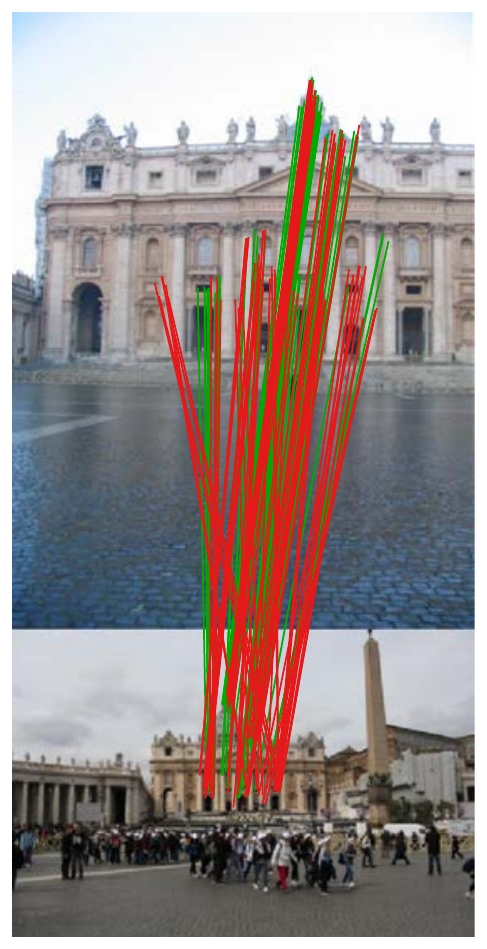}\hspace{\imsp} &
\includegraphics[height=\imh]{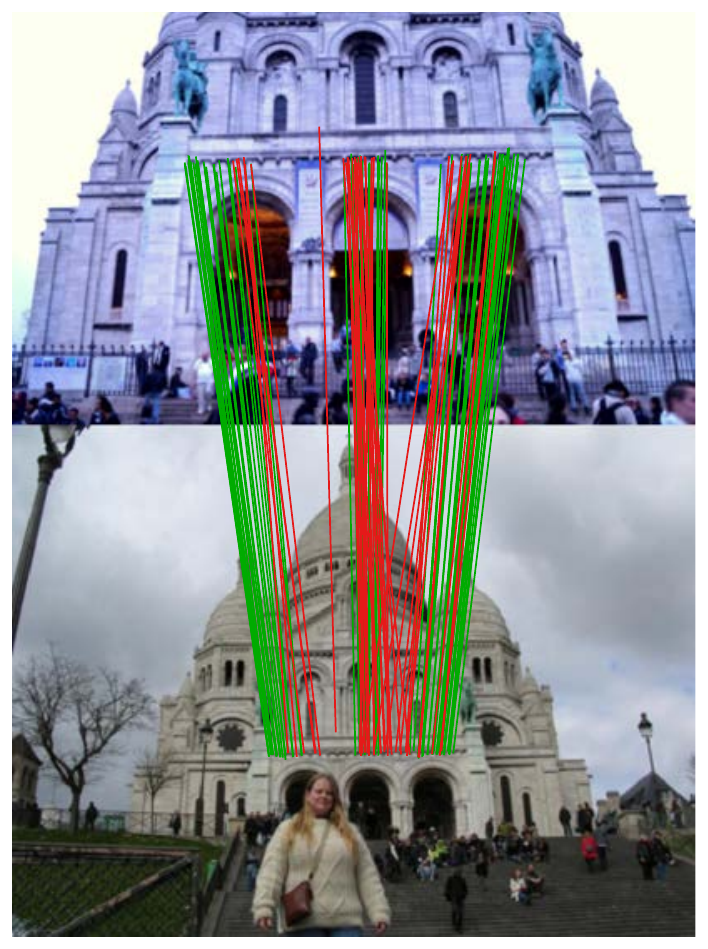}\hspace{\imsp} &
\includegraphics[height=\imh]{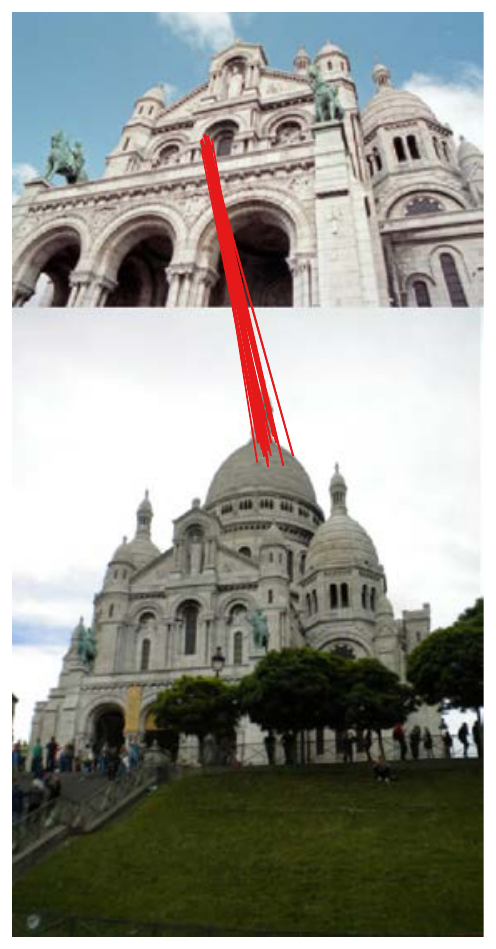}\hspace{\imsp} &
\includegraphics[height=\imh]{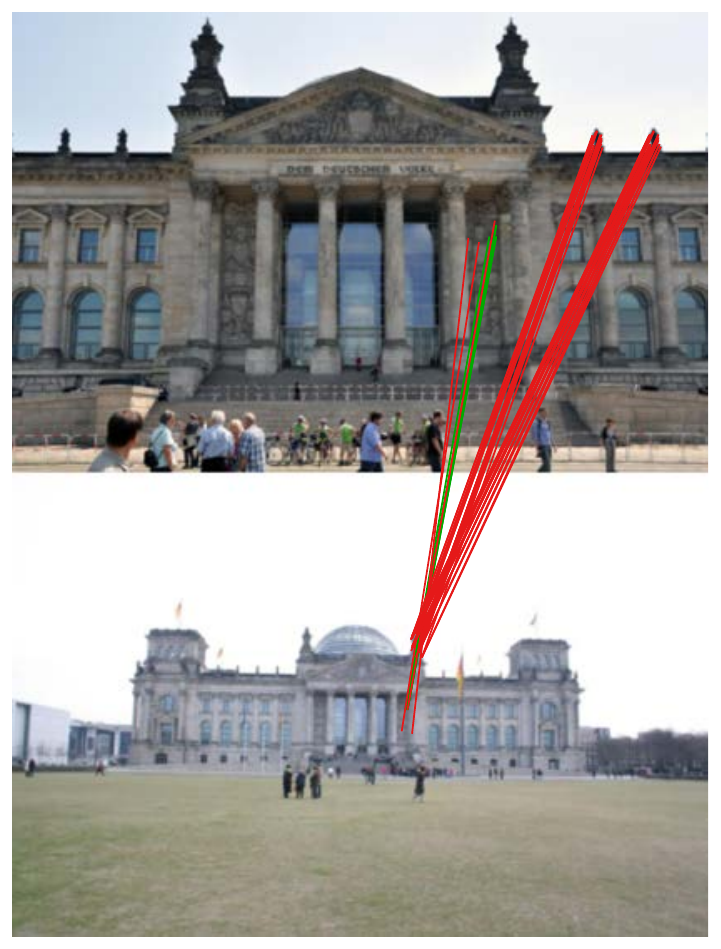}\hspace{\imsp} &
\includegraphics[height=\imh]{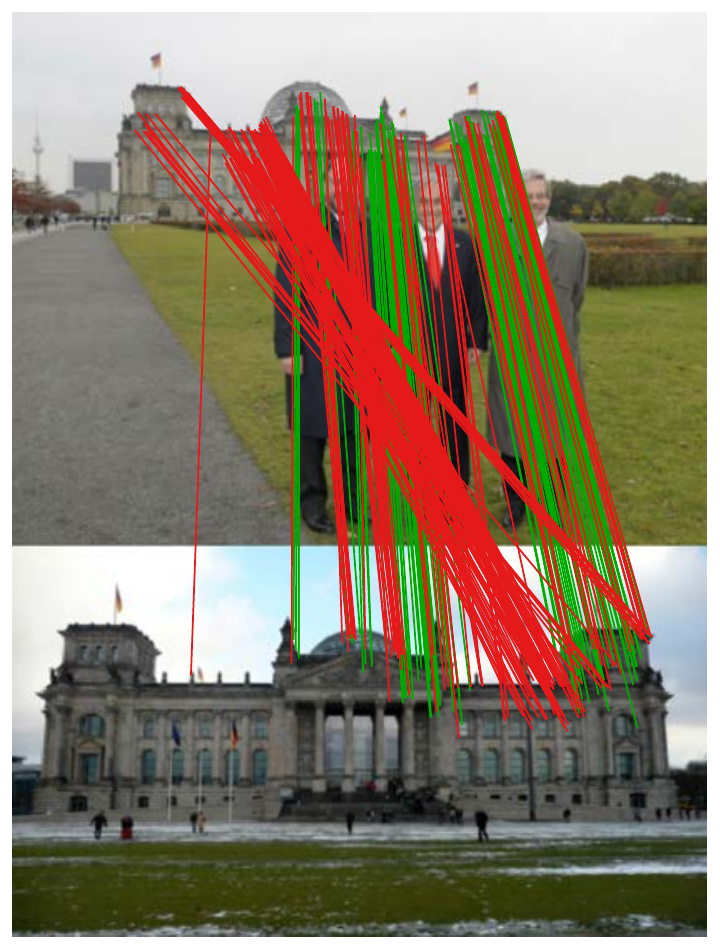}\hspace{\imsp} &
\includegraphics[height=\imh]{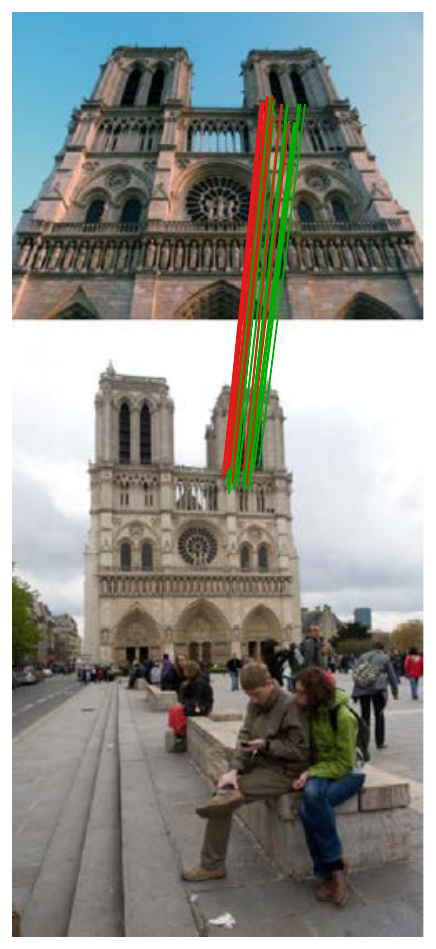}\hspace{\imsp} &
\includegraphics[height=\imh]{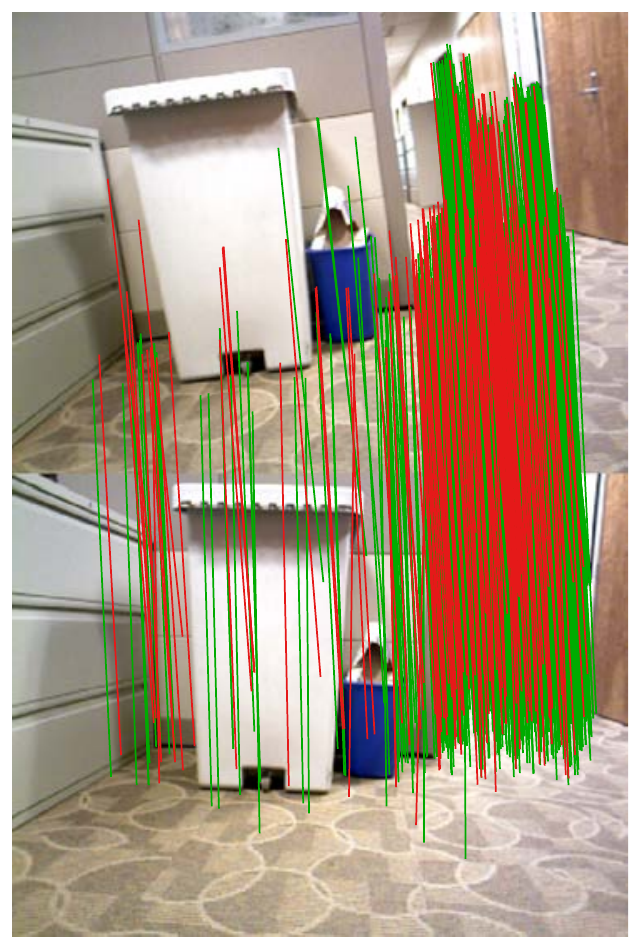}\hspace{\imsp} &
\includegraphics[height=\imh]{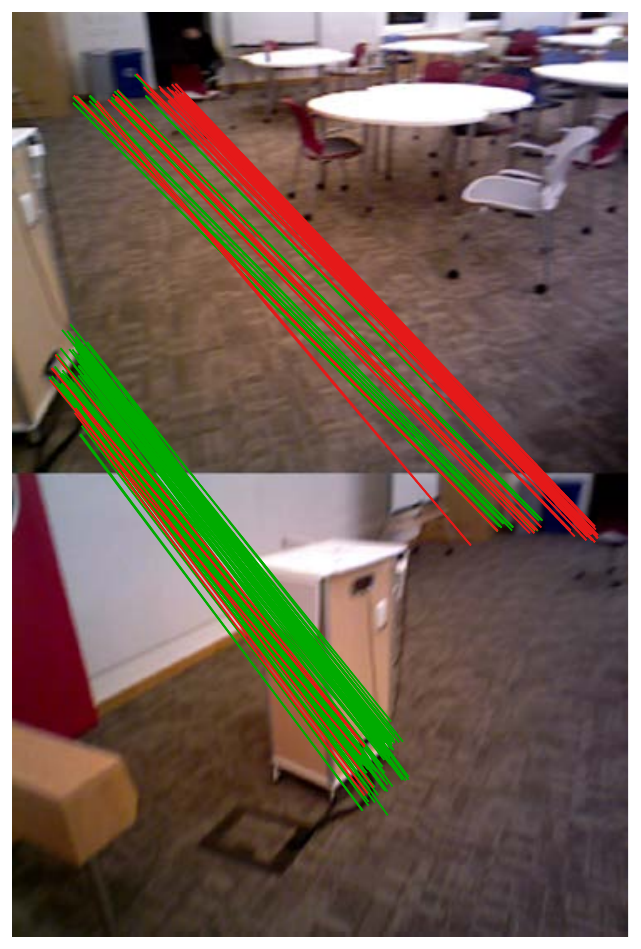}\hspace{\imsp} &
\includegraphics[height=\imh]{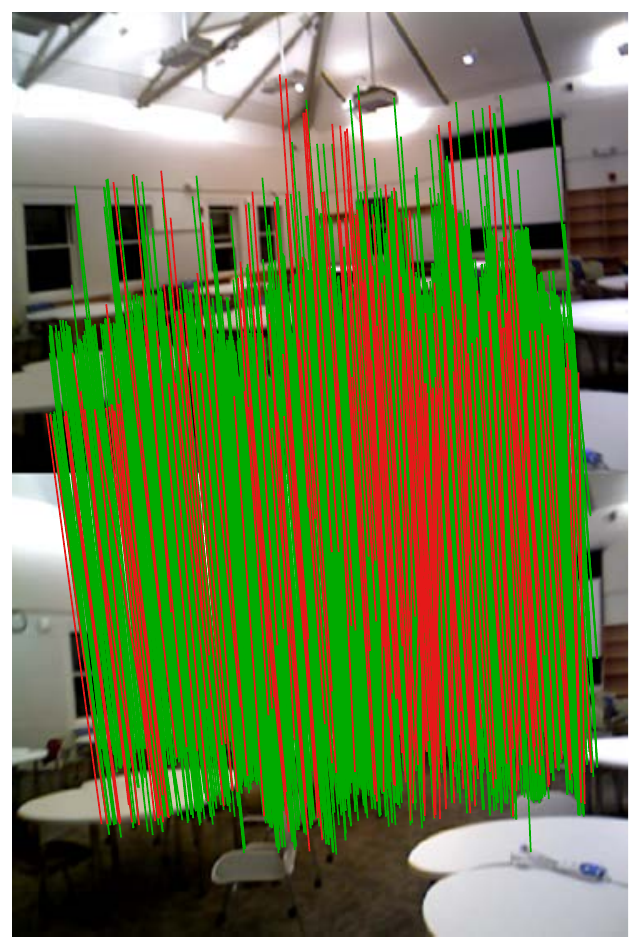}\hspace{\imsp} \\
\includegraphics[height=\imh]{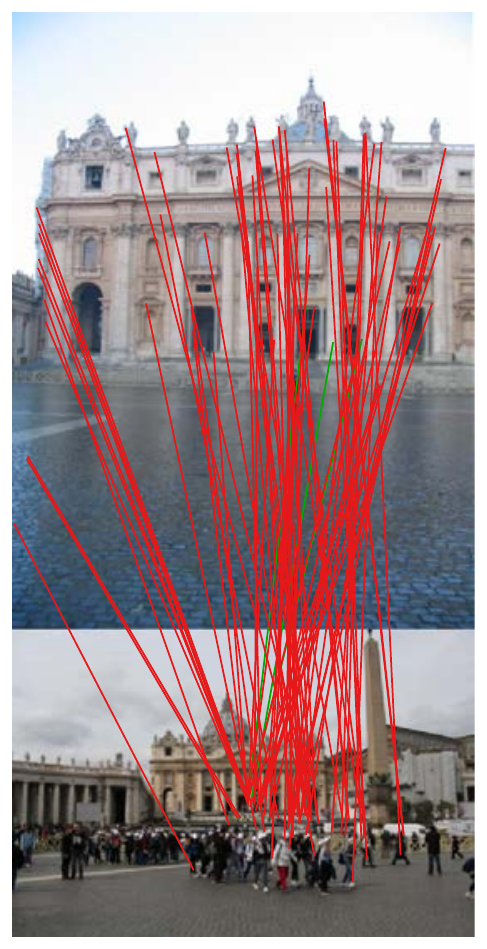}\hspace{\imsp} &
\includegraphics[height=\imh]{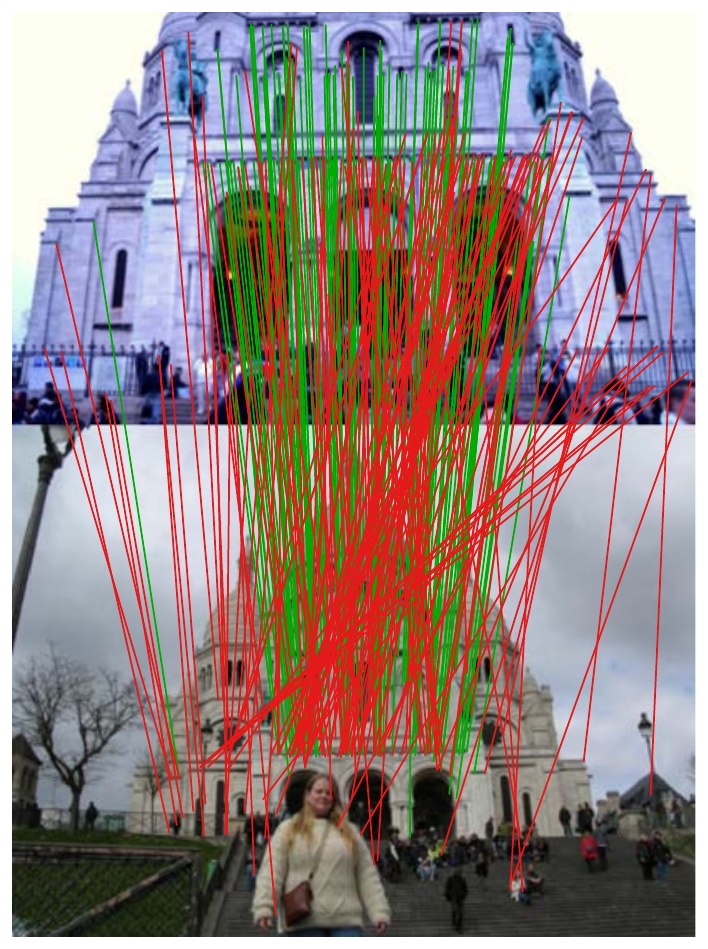}\hspace{\imsp} &
\includegraphics[height=\imh]{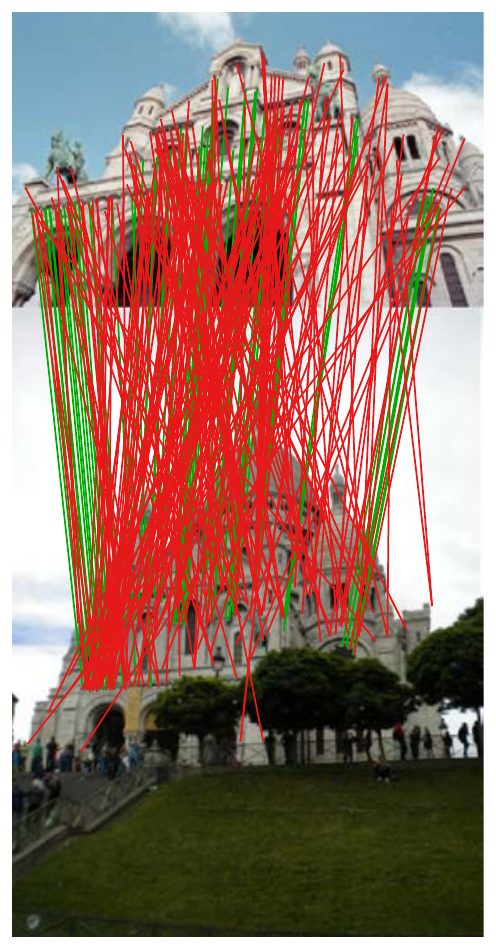}\hspace{\imsp} &
\includegraphics[height=\imh]{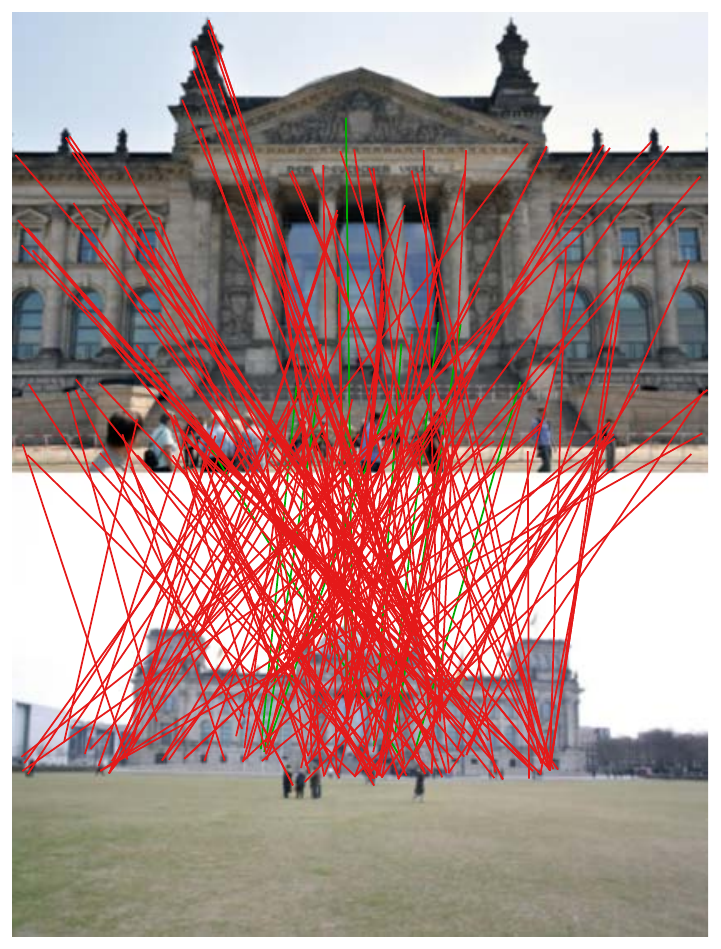}\hspace{\imsp} &
\includegraphics[height=\imh]{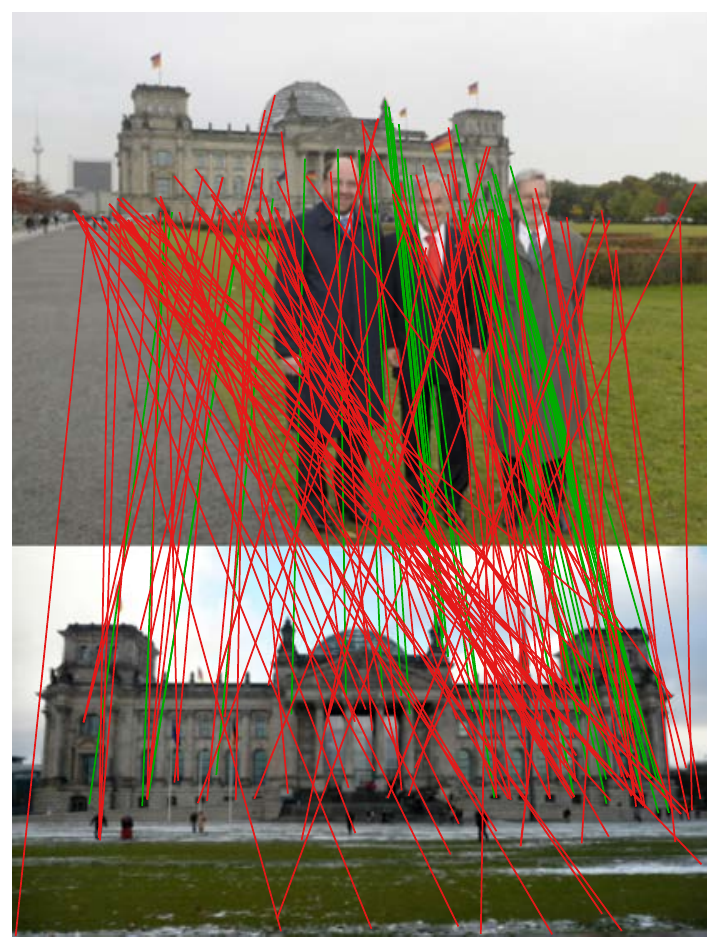}\hspace{\imsp} &
\includegraphics[height=\imh]{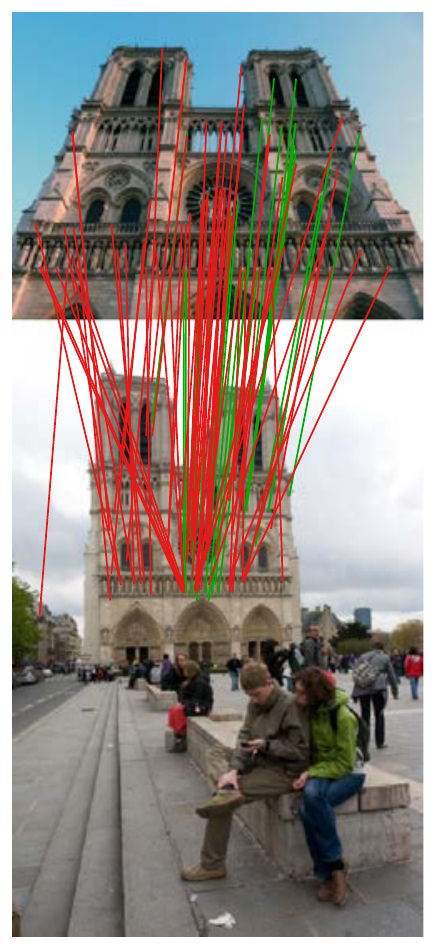}\hspace{\imsp} &
\includegraphics[height=\imh]{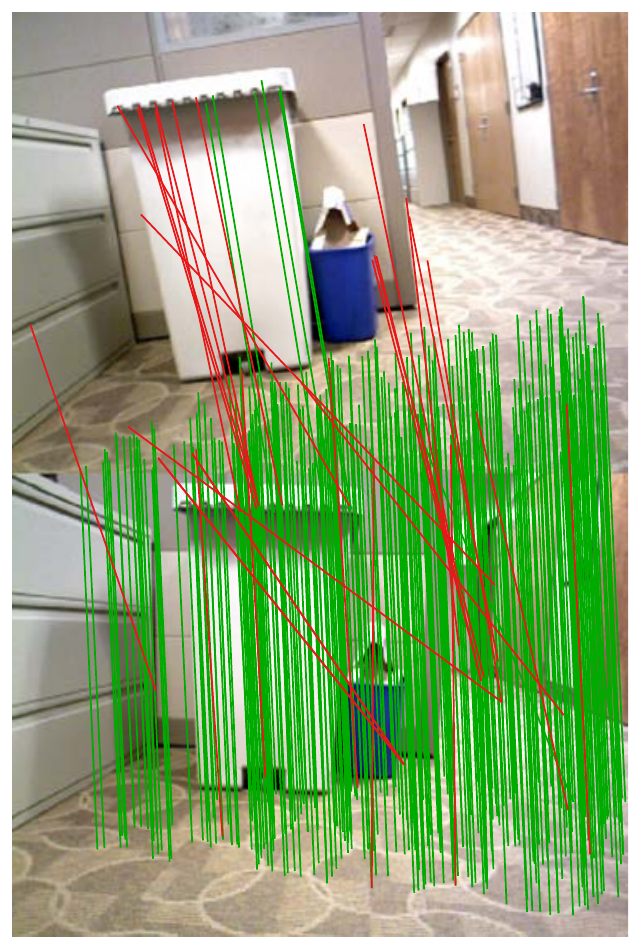}\hspace{\imsp} &
\includegraphics[height=\imh]{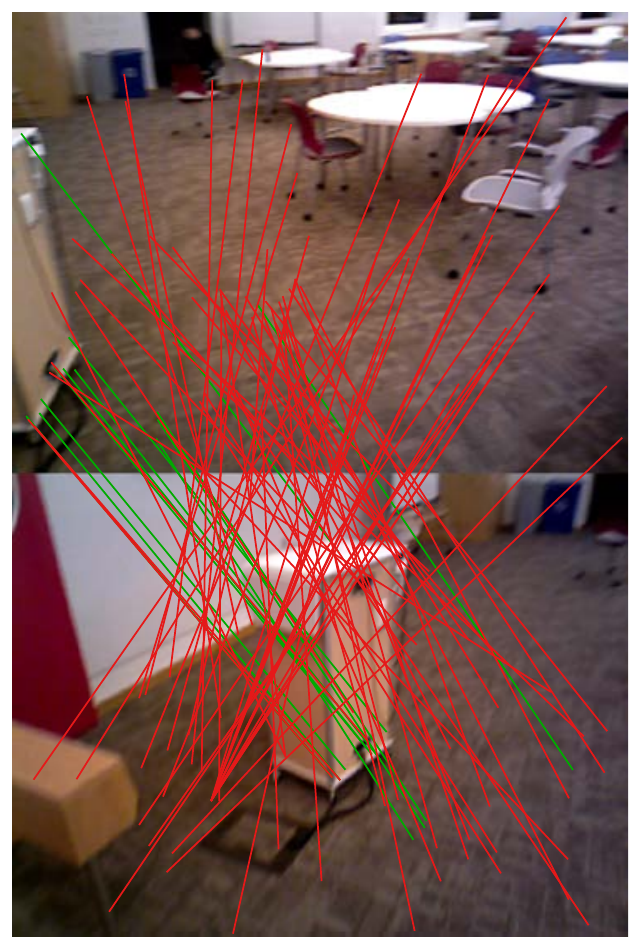}\hspace{\imsp} &
\includegraphics[height=\imh]{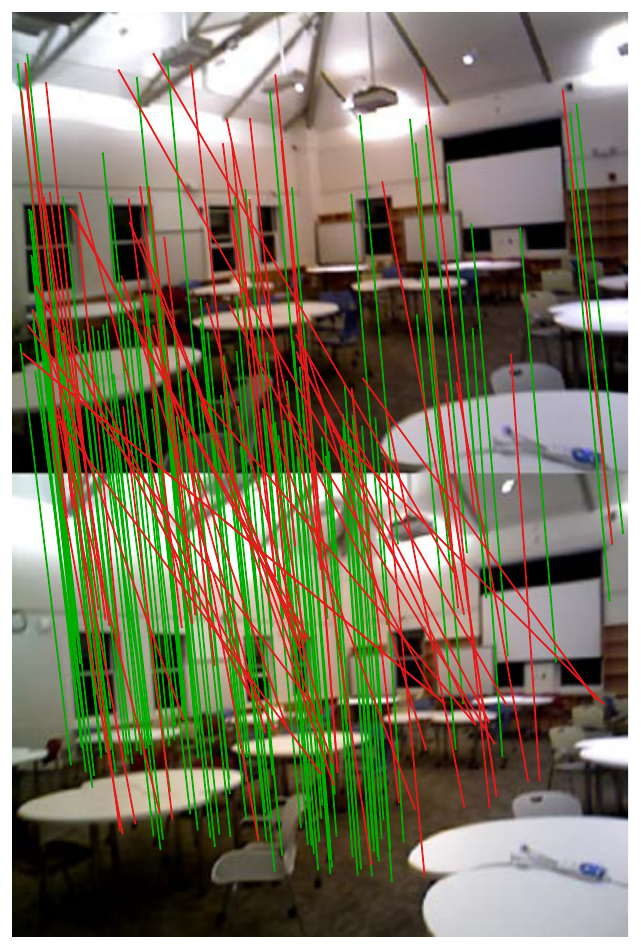}\hspace{\imsp} \\
\includegraphics[height=\imh]{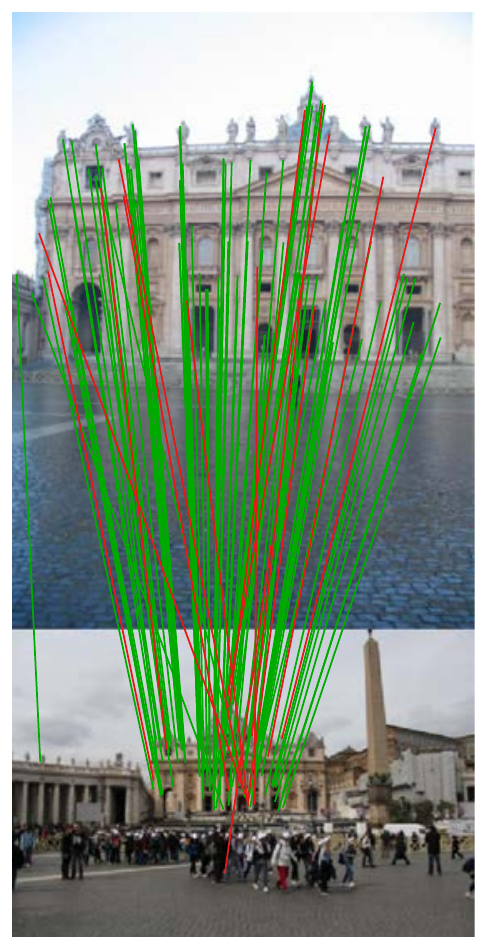}\hspace{\imsp} &
\includegraphics[height=\imh]{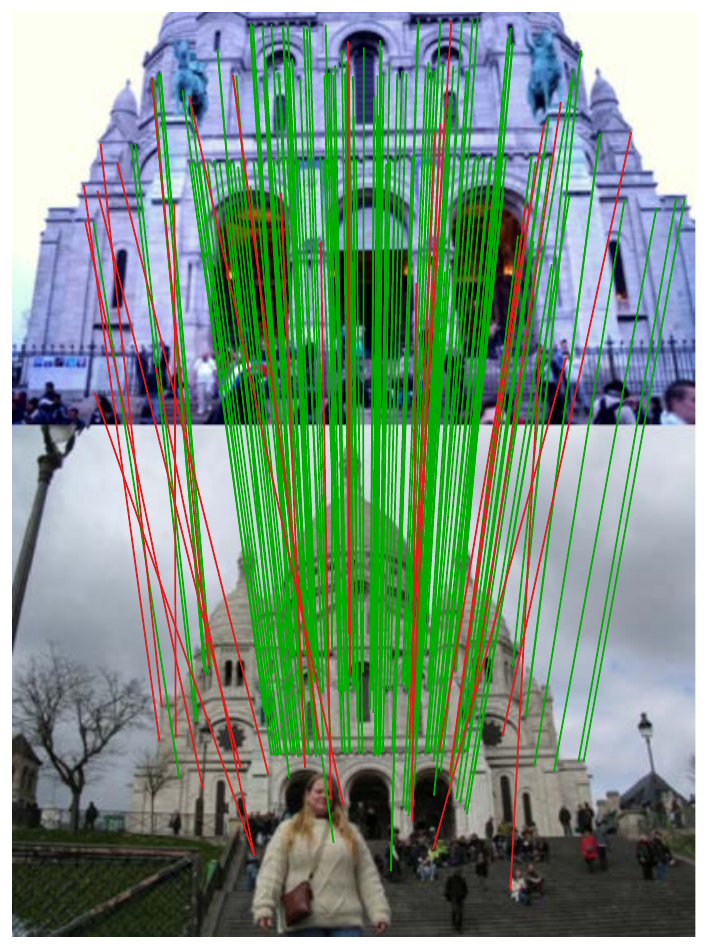}\hspace{\imsp} &
\includegraphics[height=\imh]{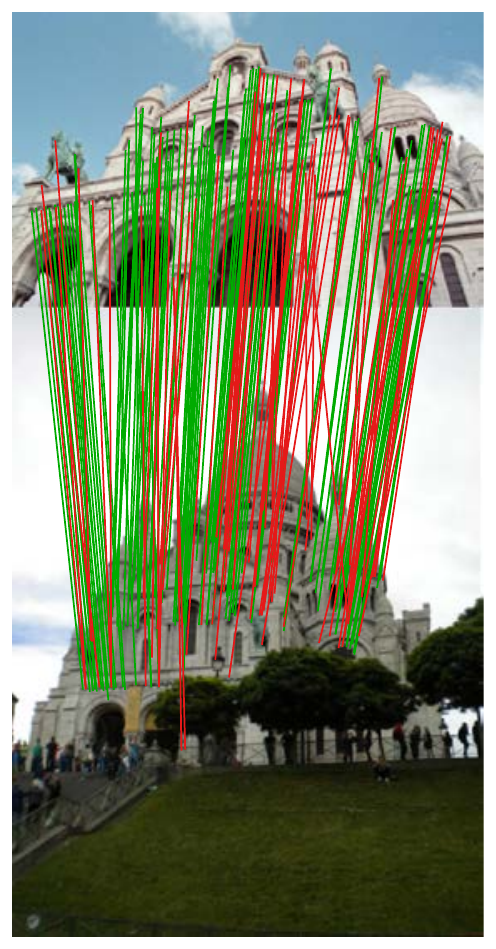}\hspace{\imsp} &
\includegraphics[height=\imh]{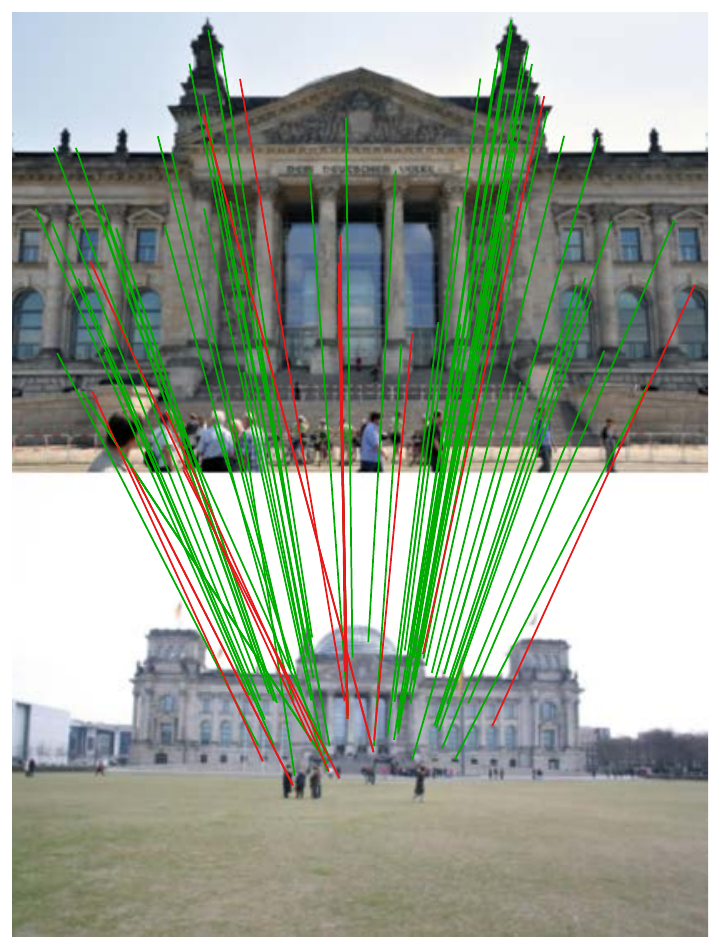}\hspace{\imsp} &
\includegraphics[height=\imh]{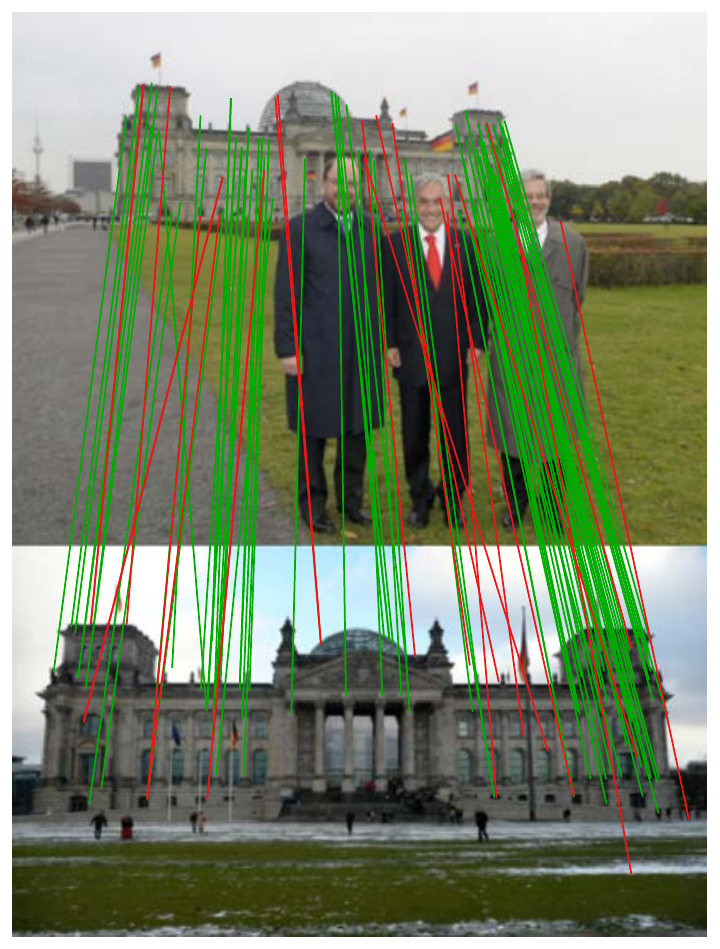}\hspace{\imsp} &
\includegraphics[height=\imh]{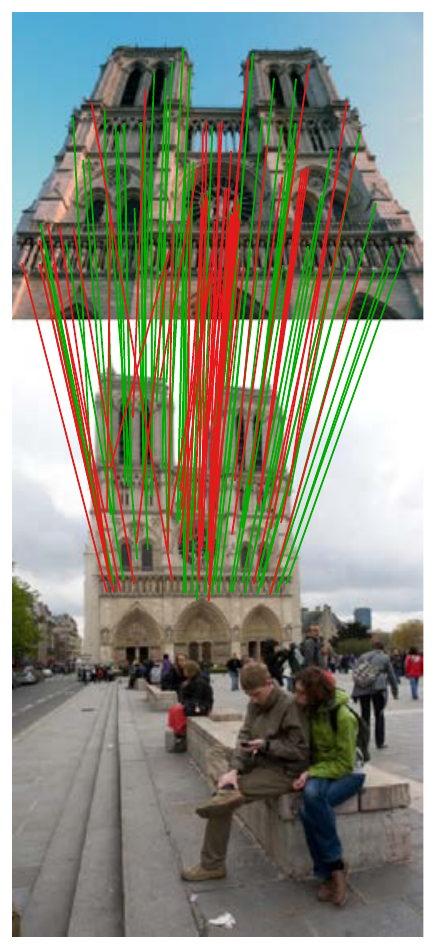}\hspace{\imsp} &
\includegraphics[height=\imh]{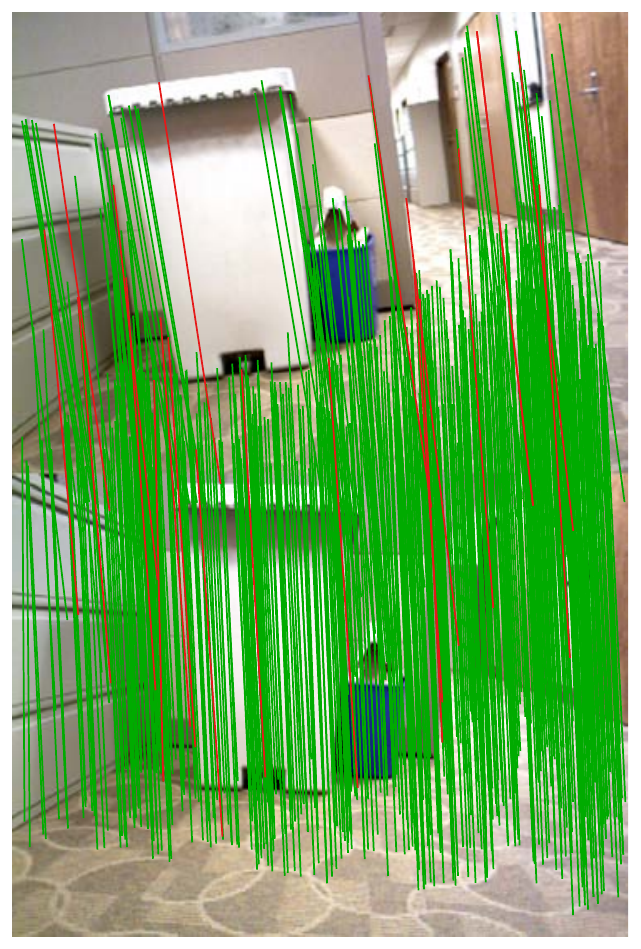}\hspace{\imsp} &
\includegraphics[height=\imh]{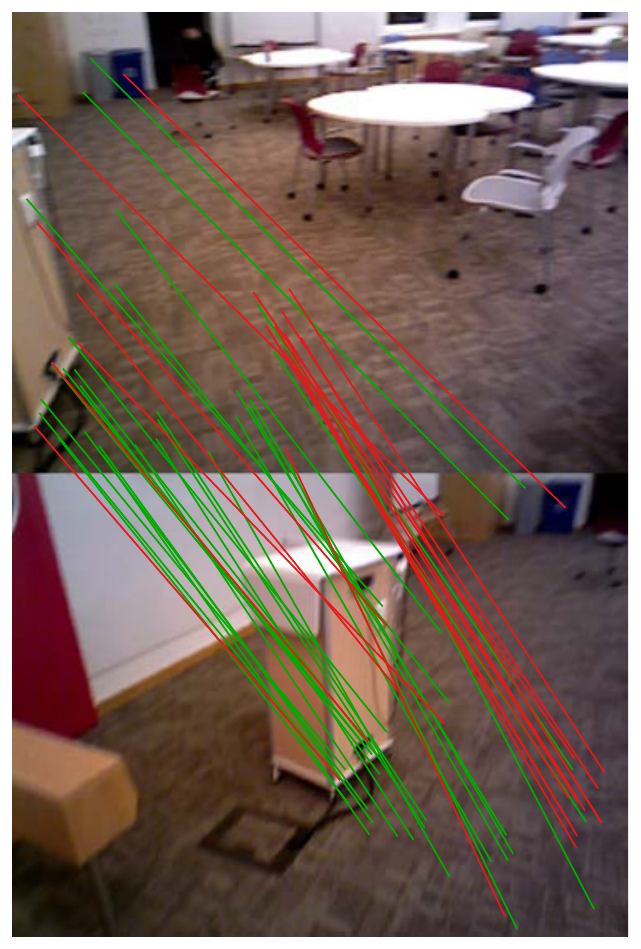}\hspace{\imsp} &
\includegraphics[height=\imh]{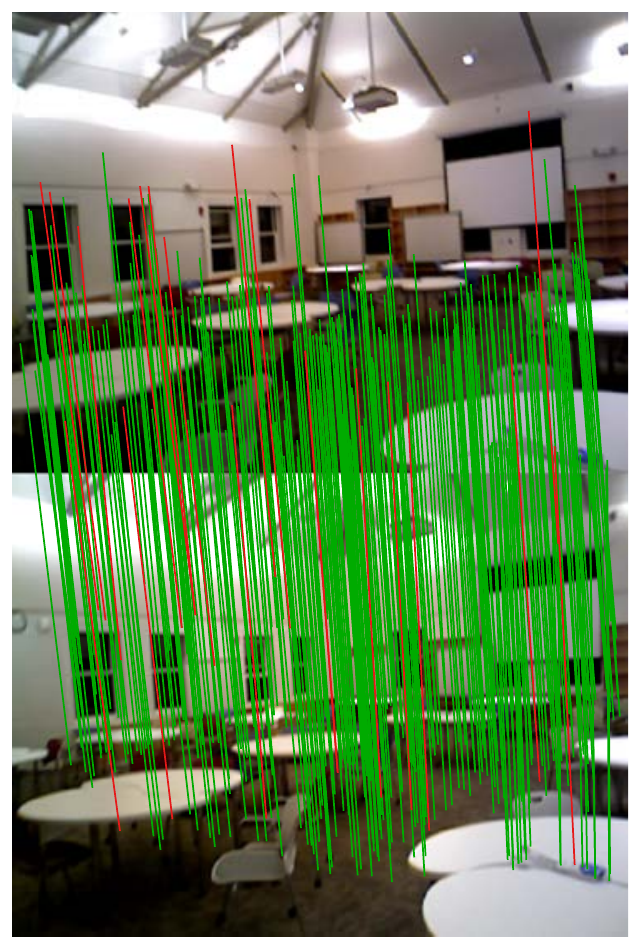}\hspace{\imsp}\\
\end{tabular}
\caption{
  Matches from ({\bf top}) GMS with $10$k ORB features, ({\bf middle}) RANSAC,
  and ({\bf bottom}) our approach, with the same $2$k SIFT features. Matches are
  in green if their symmetric epipolar distance in normalized coordinates is
  below 0.01, and red otherwise.}
\label{fig:qualitative}
\vspace{-3mm}
\end{figure*}

\subsection{Ablation Study}
\label{sec:choices} 

At the heart of our approach are two key ideas: to label the correspondences as
inliers or outliers while simultaneously using them to estimate the essential
matrix, and to leverage a PointNet-like architecture, replacing its global
context feature with our Context Normalization, introduced in
Section~\ref{sec:network}. We examine the impact of these two choices.

\subsubsection{Hybrid Approach vs Classification vs Regression}
\label{sec:variants}

To show the benefits of our hybrid approach, we compare four different settings
of our algorithm:
\begin{itemize}
\itemsep0em

\item {\bf Ours:} Our complete formulation with $\alpha$ and $\beta$ in \eq{hybrid_loss} set to $1$ and $0$ initially and then to $1$ and $0.1$ after $20$k batches. In other words, we first seek to assign reasonable weights to the correspondences before also trying to make the essential matrix accurate.
 
\item {\bf Essential:} We disable the classification loss by setting $\alpha=0$ and $\beta=1$ in \eq{hybrid_loss}. This amounts to direct regression from correspondences to the essential matrix by assigning weights to the correspondences and performing least-squares fitting.

\item {\bf Classification:} We disable the essential loss, setting $\alpha=1$
  and $\beta=0$ in \eq{hybrid_loss}. The network then tries to classify
  correspondences into inliers and outliers.
 
\item {\bf Direct:} We regress the essential matrix directly by average-pooling the output of our last ResNet block and adding a fully-connected layer that outputs nine values, which we take to be the output of $g$ in \eq{essential_loss}. In other words, we directly predict the coefficients of the essential matrix without resorting to a weighted version of the 8-point algorithm, as in {\bf Essential}.
\end{itemize}
%
We ran these four variants on the `Sacre Coeur' collection from YFCC100M, using
SIFT keypoints, and report the results in \fig{results_loss}.  {\bf Essential}
and {\bf Direct}, which do direct regression without
classification, perform worse than {\bf Classification}, as long as the number of 
keypoints is sufficient. {\bf Ours} outperforms all the others by combining
both classification and regression, by a margin of 12-24\%. Note that the difference is larger
for smaller error thresholds, suggesting that
both {\bf Essential} and {\bf Direct} are learning the general trend of
the dataset without providing truly accurate poses.


\vspace{-3mm}
\subsubsection{Context Normalization vs Context Feature}
\label{sec:results_architecture}

For comparison purposes, we reformulate our approach using the PointNet
architecture~\cite{Qi17}.
The specificity of PointNet is that it extracts a global context feature from
the input features, which is then concatenated to each individual feature and
re-injected into the network.
By contrast, we simply embed it into each point-wise feature through
Context Normalization.
For a fair comparison, we use an architecture of
similar complexity to ours, with 12 MLPs each for the extraction of point
features, global features, and the final output. We replace max pooling by
average pooling to extract the global feature, which gave better results. We
refer to this architecture as {\bf PointNet}, and train it using the
hybrid loss of Eq.~\eqref{eq:hybrid_loss}, as we do for {\bf
  Ours}. For completeness, we also try our approach without Context
Normalization.

We report results on the `Sacre Coeur' sequence in \fig{results_architecture}.
In all three cases, we present results without and with the final RANSAC stage
of \secref{ransac}. As expected, our approach without Context Normalization
performs poorly, whereas, with it, it does much better than {\bf PointNet}.

\subsection{Post-processing with RANSAC}
\label{sec:results_ransac}

Note that traditional keypoint-based methods do not work at all without
RANSAC. By contrast, our network outperforms RANSAC in a {\em single forward
pass}, using only the 8-point algorithm. However, at test time we can drop
the differentiability constraints, as explained in \secref{ransac}, and apply
RANSAC on our surviving inliers to further boost performance. We do this for the
remainder of this paper.

Moreover, our method is much faster than stand-alone RANSAC because it
can throw away most bad matches in a single step. Given $2$k matches, a
forward pass through our network takes 13 ms on GPU (or 25 ms on CPU) and
returns on average \texttildelow{}300 inliers---RANSAC then removes a further
\texttildelow{}100 matches in 9 ms. By contrast, a RANSAC loop with the full
$2$k matches needs 373 ms and returns \texttildelow{}300 inliers. Our method,
including RANSAC, is thus not only more accurate but also 17x faster (please
refer to the supplementary material for comprehensive numbers).
%
%

\subsection{Comparison to the Baselines}
\label{sec:comparison}

We now turn to comparing our method against the baselines discussed above.
Given the results presented in \secref{choices}, we use {\bf Ours} with context
normalization turned on, and apply RANSAC post-processing to {\it all} of
the sparse methods. We average the results over multiple sets---please refer to
the supplementary material for additional results.

We first consider networks trained and
tested on images from the same scene, and then on
different scenes. Note that we split the images into disjoint sets for training
an testing, as explained in Section~\ref{sec:datasets}, so there is no overlap
between both sets in either case. For keypoint-based methods other than GMS, we
consider both SIFT and LIFT.

\vspace{-3mm}
\subsubsection{Performance on Known Scenes}
\label{sec:results_sameset}

Consider the 5 collections of outdoor images from YFCC100M and the 9 indoor
sequences from SUN3D, described in Section~\ref{sec:datasets} and depicted by
\fig{pair-examples}. We report our comparative results in
\figs{results_sameset}{qualitative}. The training and test sets
are always disjoint, but drawn from the {\it same} collection.

For the five `Outdoors' collections, whose images are feature-rich, we achieve
our best results using LIFT features trained on photo-tourism datasets.  {\bf
  Ours} then delivers an mAP that is more than twice that of previous
state-of-the-art methods. However, even when using the more popular SIFT, it
still outperforms the other methods. Furthermore, the gap grows wider for small
error thresholds, indicating that our approach performs better the more strict
we are.

By contrast, the two `Indoors' collections feature poorly textured images with
repetitive patterns, static environments, and consistent scales, which makes
them ill-suited for keypoint-based methods and more amenable to dense
methods. Nevertheless, our method significantly outperforms all compared
methods, including dense ones.  On these images, {\bf Ours} performs better
with SIFT than LIFT, which is consistent with the fact that LIFT was trained on
photo-tourism images.  Note that for the numbers reported in
\fig{results_sameset} for DeMoN, we used their pre-trained models as it is not
possible to re-train it due to the lack of depth information for the `Outdoors'
data. Note also that their training sets include not one but many sequences
from SUN3D~\cite{Ummenhofer17}.

\vspace{-3mm}
\subsubsection{Generalization to Unknown Scenes}
\label{sec:results_generalization}

Here, we evaluate the generalization capability of our method by training and
testing on different scenes.  \fig{results_generalization} reports results for
the model trained with the combination of the `Saint Peter's' sequence from
`Outdoors' and the `Brown 1' sequence from `Indoors'.  For `Outdoors' we
report the average result for all sets excluding `Saint Peter's'. For
`Indoors', we report the average result for the 15 test-only sequences selected
by~\cite{Ummenhofer17}, for fair comparison.

We outperform all baselines on `Outdoors' by a significant
margin.
Comparing \figs{results_sameset}{results_generalization} shows that
our models generalize very well to unknown scenes.  Note that the jump in the
performance of several baselines between
\figs{results_sameset}{results_generalization} is solely due to the addition of
two easy sequences, `Fountain' and `Herzjesu', for testing in
\fig{results_generalization}.

On `Indoors' we outperform the state of the art by a small margin and lose some
generalization power, probably limited by the capabilities of SIFT and LIFT for
indoor scenes.  Nevertheless, we outperform the state of the art on {\em
  both} subsets with {\em a single model} trained on only 2000 outdoors and 500
indoors images and tested on completely different scenes.
Further results are given in the supplementary material, where we show that we
can still outperform the state of the art with much smaller training sets,
\eg, 59 images.

\vspace{-3mm}




\section{Conclusion}

We have proposed a single-shot technique to recover the relative motion between
two images with a deep network. In contrast with current trends, our method is
sparse, indicating that keypoint-based robust estimation can still be relevant
in the age of dense, black-box formulations.  Our approach outperforms the
state of the art by a significant margin with few images and limited
supervision.

Our solution requires known intrinsics.  In the future, we plan to investigate
using the fundamental matrix instead of the essential. While the formalism
would remain largely unchanged, we expect numerical stability problems in the
regression component of the hybrid loss, which may require additional
normalization layers or regularization terms.




{\small
\bibliographystyle{ieee}
\bibliography{short,vision,learning}
}

\cleardoublepage
\onecolumn
\appendix 

\section{Supplementary Appendix for ``Learning to Find Good Correspondences''}
\label{sec:appendix}

\subsection{Dataset details}
\label{sec:appendix_datasets}

Here we detail the sequences used for training and testing in
\tbl{supp_datasets}. The image numbers reported for SUN3D~\cite{Xiao13} are
after subsampling the video sequences by a factor of 10. For SUN3D we choose 9
sequences for training, and use the 15 sequences previously used
by~\cite{Ummenhofer17}, collectively marked with $\ddagger$, only for testing.
We generate disjoint training, validation and test subsets by splitting the
images in each set in a 60-20-20 ratio, as explained in \secref{datasets}. For
$\ddagger$ we take up to 500 images with a 0-0-100 ratio, as we do not train
any model on them.

The last column assigns a label (a-u) to each set for convenience, which is
used in \secref{appendix_perseq}. Our best model is trained concatenating the
datasets marked with $\Diamond$ and $\triangle$.

\begin{table}[!h]
  \begin{center}
\small
\begin{tabular}{lcccc}
  \toprule
  Scene & Images & 3D points & Avg. views & Label \\
  \midrule
  \multicolumn{4}{c}{Yahoo YFCC100M~\cite{Thomee16,Heinly15}} \\
  \midrule
  `Buckingham'    & 1676 & 152003 & 11.83 & a \\
  `Notre Dame'    & 3767 & 502017 & 35.41 & b\\
  `Sacre Coeur'   & 1179 & 152594 & 19.63 & c \\
  `Saint Peter's' & 2506 & 235668 & 26.96 & $\Diamond$ \\
  `Reichstag'     & 75 & 19881 & 7.74 & d \\
  \midrule
  \multicolumn{4}{c}{Multi-View Stereo~\cite{Strecha08b}} \\
  \midrule
  `Fountain'    & 11 & -- & -- & e \\
  `HerzJesu'    & 8 & -- & -- & f \\
  \midrule
  \multicolumn{4}{c}{SUN3D~\cite{Xiao13} (training and validation)} \\
  \midrule
  `Harvard 1' {\scriptsize (\texttt{harvard\_conf\_big/hv\_conf\_big\_1})} & 455 & -- & -- & -- \\
  `Harvard 2' {\scriptsize (\texttt{harvard\_computer\_lab/hv\_c1\_1})} & 543 & -- & -- & -- \\
  `Harvard 3' {\scriptsize (\texttt{harvard\_corridor\_lounge/hv\_lounge\_corridor2\_1})} & 540 & -- & -- & -- \\
  `Harvard 4' {\scriptsize (\texttt{harvard\_corridor\_lounge/hv\_lounge\_corridor3\_whole\_floor})} & 629 & -- & -- & -- \\
  `Brown 1' {\scriptsize (\texttt{brown\_bm\_3/brown\_bm\_3})} & 841 & -- & -- & $\triangle$ \\
  `Brown 2' {\scriptsize (\texttt{brown\_cs\_4/brown\_cs4})} & 877 & -- & -- & -- \\
  `Hotel 1' {\scriptsize (\texttt{hotel\_ucla\_ant/hotel\_room\_ucla\_scan1\_2012\_oct\_05})} & 1305 & -- & -- & -- \\
  `Hotel 2' {\scriptsize (\texttt{hotel\_pedraza/hotel\_room\_pedraza\_2012\_nov\_25})} & 1065 & -- & -- & -- \\
  `Home' {\scriptsize (\texttt{home\_pt/home\_pt\_scan1\_2012\_oct\_19})} & 2407 & -- & -- & -- \\
  \midrule
  \multicolumn{4}{c}{SUN3D~\cite{Xiao13} (test only, chosen by~\cite{Ummenhofer17}) ($\ddagger$)} \\
  \midrule
  {\scriptsize \texttt{brown\_cogsci\_2/brown\_cogsci\_2}} & 259 & -- & -- & g \\
  {\scriptsize \texttt{brown\_cogsci\_6/brown\_cogsci\_6}} & 500 & -- & -- & h \\
  {\scriptsize \texttt{brown\_cogsci\_8/brown\_cogsci\_8}} & 126 & -- & -- & i \\
  {\scriptsize \texttt{brown\_cs\_3/brown\_cs3}} & 340 & -- & -- & j \\
  {\scriptsize \texttt{brown\_cs\_7/brown\_cs7}} & 251 & -- & -- & k \\
  {\scriptsize \texttt{hotel\_florence\_jx/florence\_hotel\_stair\_room\_all}} & 500 & -- & -- & l \\
  {\scriptsize \texttt{harvard\_c4/hv\_c4\_1}} & 224 & -- & -- & m \\
  {\scriptsize \texttt{harvard\_c10/hv\_c10\_2}} & 81 & -- & -- & n \\
  {\scriptsize \texttt{harvard\_corridor\_lounge/hv\_lounge1\_2}} & 154 & -- & -- & o \\
  {\scriptsize \texttt{harvard\_robotics\_lab/hv\_s1\_2}} & 159 & -- & -- & p \\
  {\scriptsize \texttt{mit\_32\_g725/g725\_1}} & 377 & -- & -- & q \\
  {\scriptsize \texttt{mit\_46\_6conf/bcs\_floor6\_conf\_1}} & 327 & -- & -- & r \\
  {\scriptsize \texttt{mit\_46\_6lounge/bcs\_floor6\_long}} & 500 & -- & -- & s \\
  {\scriptsize \texttt{mit\_w85g/g\_0}} & 387 & -- & -- & t \\
  {\scriptsize \texttt{mit\_w85h/h2\_1}} & 500 & -- & -- & u \\
  \bottomrule
\end{tabular}
  \end{center}
\label{tbl:supp_datasets}
\caption{Datasets.}
\end{table}

\subsection{Training with limited data}
\label{sec:appendix_small}

Due to space constraints, the paper only reports results with our best
model, which is the concatenation of `St. Peter's' ($\Diamond$) and
`Brown 1' ($\triangle$). Here we replicate the experiments of
\secref{results_sameset} and \secref{results_generalization}, \ie, we train a model
and evaluate it first on the same
sequence and then on every other `Outdoors' sequence, respectively, but now using
only our {\em smallest} training sequence. The dataset in question is `Reichstag' (d)
from the `Outdoors' subset, which contains only 59 images for training, 8
for validation and 8 for testing. Note that after accounting for visibility
constraints, this still lets us extract over 1500 image pairs for training and
about 35 for each validation and testing.

\begin{figure}
\centering
\includegraphics[width=0.9\linewidth]{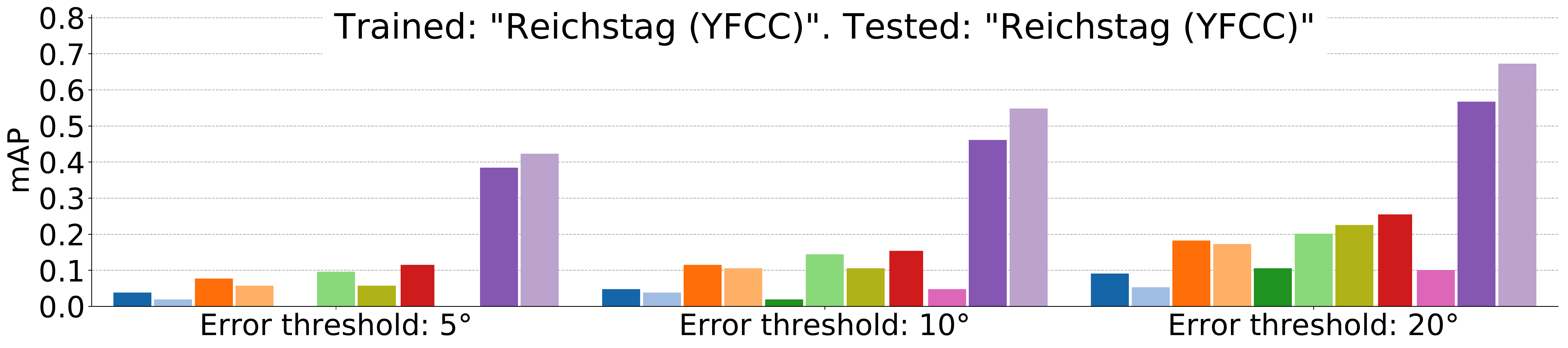} \\
\includegraphics[width=0.8\linewidth]{fig/results/legend.pdf}
\caption{Results for the model trained and tested on `Reichstag'.}
\label{fig:app_reichstag1}
\end{figure}

\begin{figure}
\centering
\includegraphics[width=0.9\linewidth]{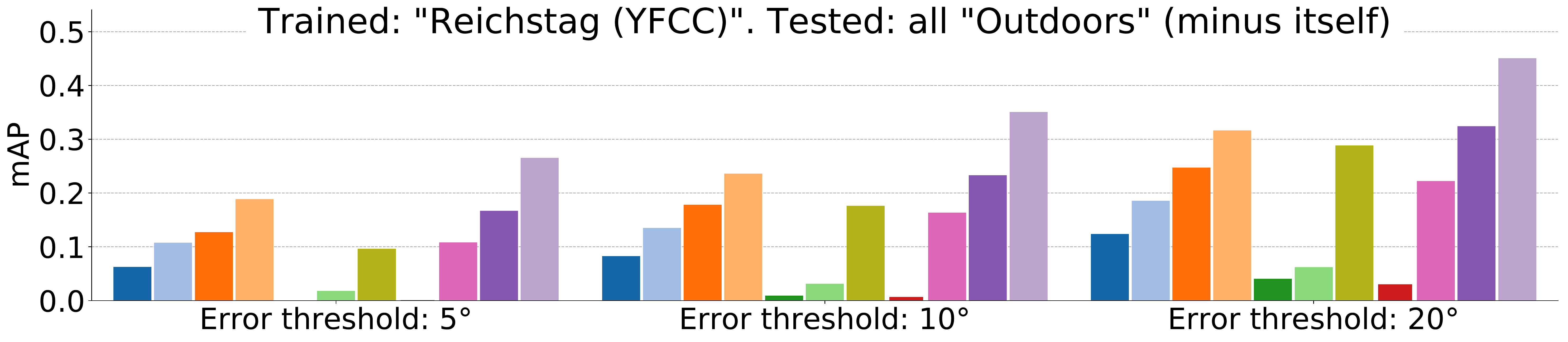} \\
\includegraphics[width=0.8\linewidth]{fig/results/legend.pdf}
\caption{Results for the model trained on `Reichstag' (d), and tested on every
other `Outdoors' sequence, \ie, a-c, e, f and $\Diamond$. We average the results
over each sequence.}
\label{fig:app_reichstag2}
\end{figure}

\fig{app_reichstag1} shows results training and testing on (different subsets of) the same sequence,
and \fig{app_reichstag2} shows how the model generalizes over every
`Outdoors' sequence other than itself, \ie, a-c, e, f, and $\Diamond$. We follow the same protocols as in
\secref{results_sameset} and \secref{results_generalization}. The best results are
obtained with LIFT features, which is consistent with our previous observations. 
Our method outperforms all the baselines, with LIFT plus RANSAC and GMS plus RANSAC
being the closest competitors. More importantly, when generalizing to other scenes
with so little training data (\fig{app_reichstag2}) we still outperform GMS by 65\%-200\% relative at different
error thresholds,
and RANSAC by about 50\% relative.

\subsection{Per-sequence results}
\label{sec:appendix_perseq}

We could not show per-sequence results in the paper due to space constraints.
\fig{app_all_seqs} provides separate results for every testing sequence
for our approach and for every baseline at multiple error thresholds. Again,
we use our models trained on a single sequence from each data type,
marked respectively with $\Diamond$ and $\triangle$ (we do the same for
G3DR~\cite{Zamir16}). The sequences used for testing include `Outdoors'
datasets a-f in \tbl{supp_datasets}, which are averaged in the column marked
$*$, and `Indoors' datasets g-u, which are averaged in $\ddagger$. We provide
numbers on top of the bars for $*$ and $\ddagger$. Our approach outperforms every baseline,
with LIFT performing better than SIFT on the `Outdoors' subset and the opposite for the
`Indoors' subset.

Note that DeMoN only achieves good performance for e and f in the `Outdoors'
sequences, and completely fails for YFCC100M sequences. G3DR shows even
worse performance, hinting that sparse methods are preferable when it comes to
photo-tourism datasets.

\newcommand{\plotw}{0.93}
\begin{figure*}
\centering
\includegraphics[width=\plotw\linewidth]{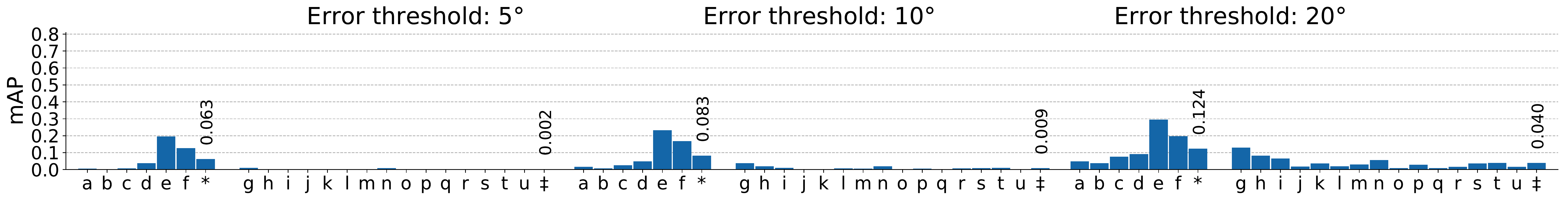} \\
\includegraphics[width=\plotw\linewidth]{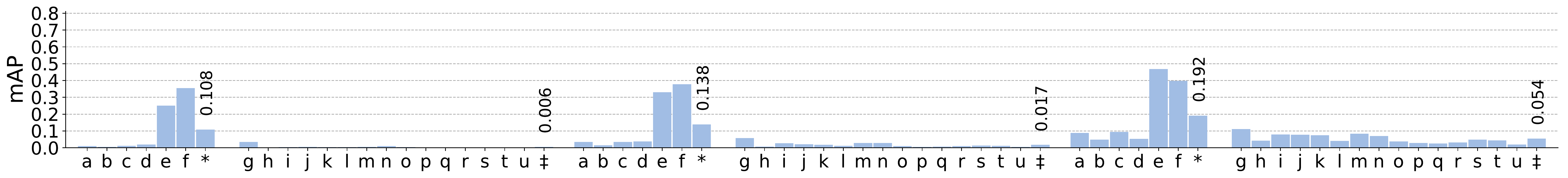} \\
\includegraphics[width=\plotw\linewidth]{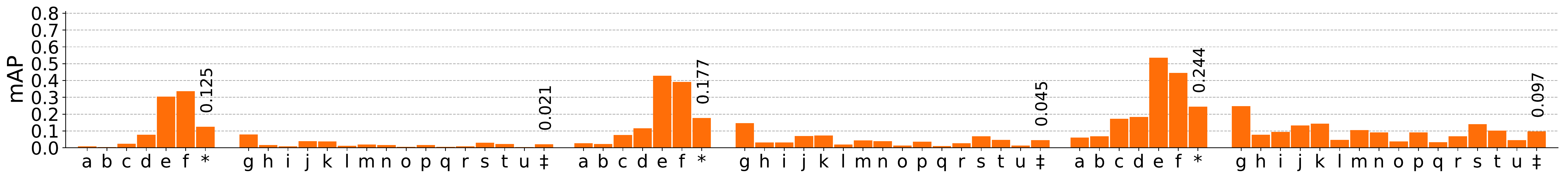} \\
\includegraphics[width=\plotw\linewidth]{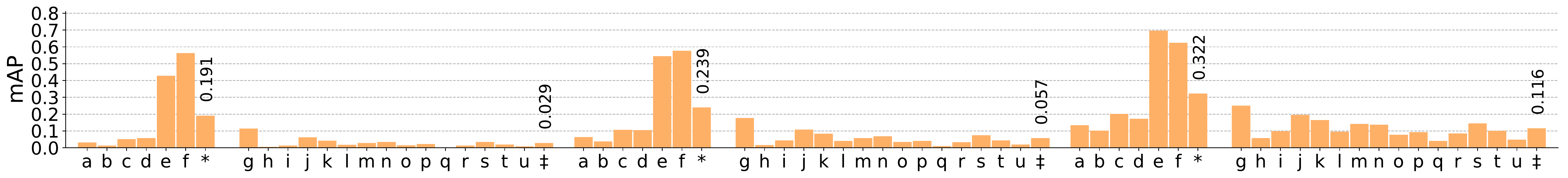} \\
\includegraphics[width=\plotw\linewidth]{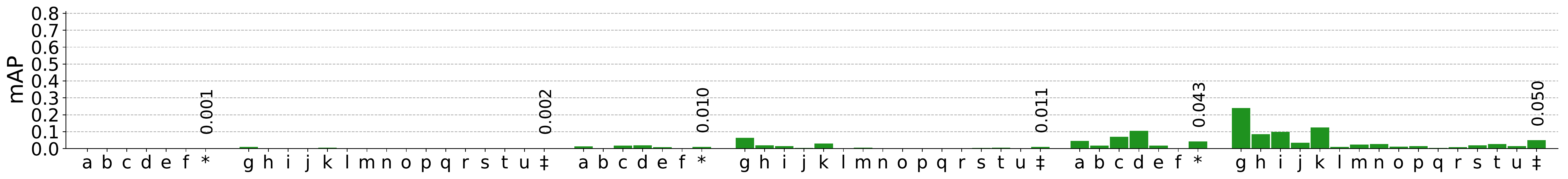} \\
\includegraphics[width=\plotw\linewidth]{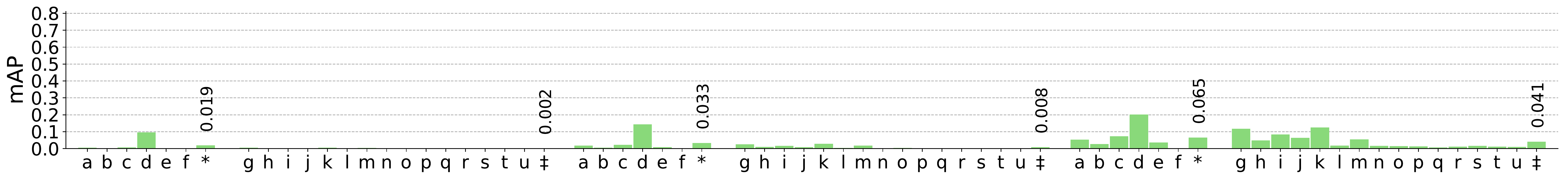} \\
\includegraphics[width=\plotw\linewidth]{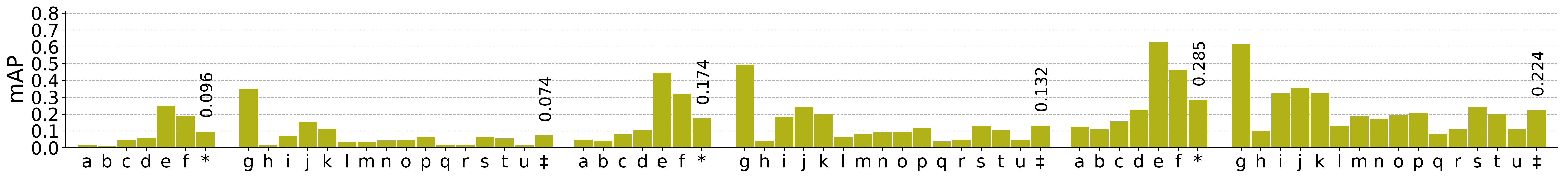} \\
\includegraphics[width=\plotw\linewidth]{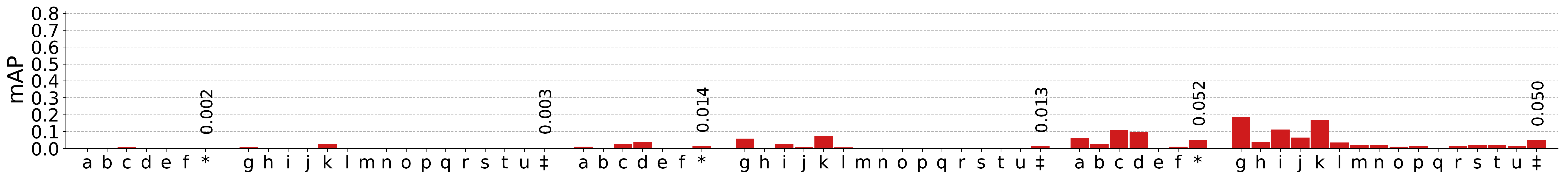} \\
\includegraphics[width=\plotw\linewidth]{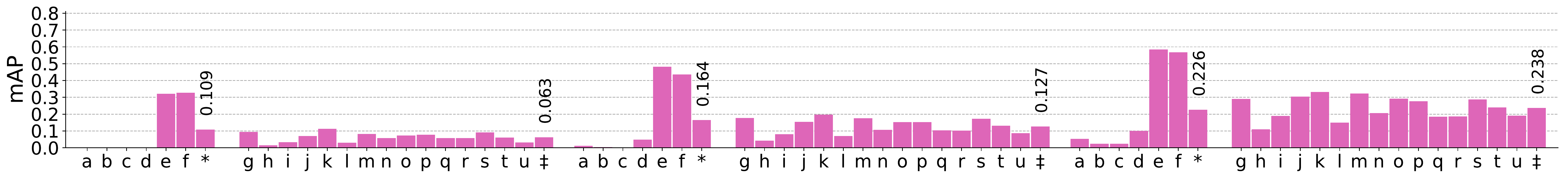} \\
\includegraphics[width=\plotw\linewidth]{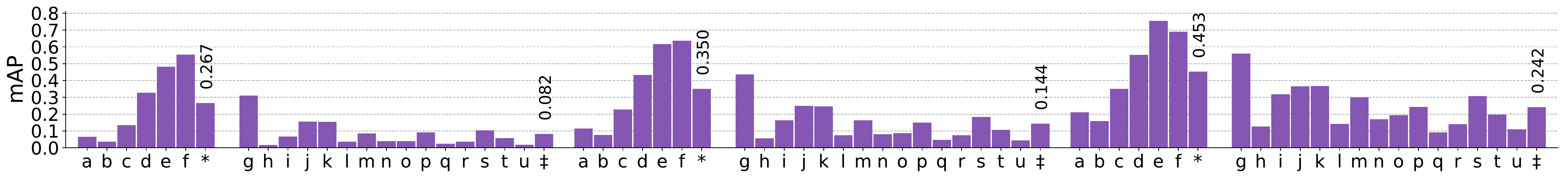} \\
\includegraphics[width=\plotw\linewidth]{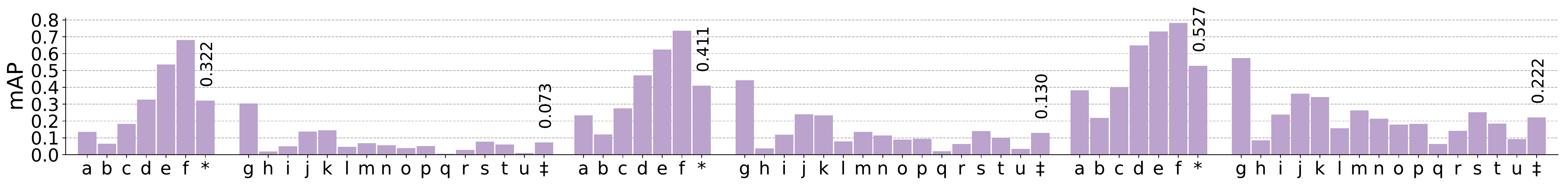} \\
\vspace{2mm}
\includegraphics[width=0.8\linewidth]{fig/results/legend.pdf}
\vspace{1mm}
\caption{Results for every sequence in the `Outdoors' subset (a-f) and the
`Indoors' subset (g-u). The entry labeled $*$ denotes the average performance
over the `Outdoors' subset, and $\ddagger$ the average performance over
the `Indoors' subset. Labels are listed in
\tbl{supp_datasets}.}
\label{fig:app_all_seqs}
\end{figure*}

\subsection{Ransac post-processing}
\label{sec:appendix_ransac}

As outlined in \ref{sec:results_ransac}, RANSAC for post-processing
allows us to greatly improve both the performance and the speed over RANSAC. In
\tbl{ransac_post} we provide results for the generalization experiments of
Section~\ref{sec:results_generalization}, for stand-alone RANSAC, and our
method using either the 8-point algorithm or RANSAC for post-processing, which were
not originally included in the paper due to spatial constraints. Note that this
boost is only possible at test time, due to the differentiability requirement
for training.

\begin{table}[!h]
\begin{center}
\small
\begin{tabular}{lcccccc}
  \toprule
  & & \multicolumn{2}{c}{Outdoors} & \multicolumn{2}{c}{Indoors} & \multirow{2}{*}{Average} \\
  & & SIFT & LIFT & SIFT & LIFT & \\
  \midrule
  RANSAC & mAP@20$^o$ & 0.221 & 0.291 & 0.097 & 0.115 & --- \\ 
  \midrule
  \multirow{2}{*}{Ours + 8-point} & mAP@20$^o$     & 0.264 & 0.343 & 0.148 & 0.143 & --- \\
                                  & w.r.t. RANSAC & +19.5\% & +17.9\% & +52.6\% & +24.3\% & +28.6\% \\
  \midrule
  \multirow{2}{*}{Ours + RANSAC} & mAP@20$^o$      & 0.462 & 0.530 & 0.242 & 0.222 & --- \\
                                  & w.r.t. RANSAC & +109.0\% & +82.1\% & +149.5\% & +93.0\% & +108.4\% \\
  \bottomrule
\end{tabular}
\end{center}
\label{tbl:ransac_post}
\caption{RANSAC vs Ours with 8-point vs Ours with RANSAC. Both SIFT and LIFT use $2$k keypoints.}
\end{table}

\subsection{Differentiating through eigendecomposition}

To differentiate through the eigendecomposition, we rely on the TensorFlow
implementation. Here, we provide a short definition for completeness. For more
details, we refer the interested readers
to~\cite{Ionescu15}. In~\cite{Ionescu15}, it is shown that for a matrix $\bX$,
which can be decomposed into $\bX = \bU \bSigma \bU^\top$, where $\bU$ is the
matrix of eigenvectors and $\bSigma$ is a diagonal matrix with eigenvalues, the
derivative w.r.t. the eigenvectors are
\begin{equation}
  d\bU = 2\bU\left(\bK \odot (\bU^\top d\bX \bU)_{sym}\right)\;,
\end{equation}
where $\bM_{sym} = \frac{1}{2}(\bM^\top + \bM)$, and
\begin{equation}
  \bK_{ij} = \begin{cases} \frac{1}{\sigma_i - \sigma_j}, & i\neq j \\
    0, & i=j \end{cases}\;,
\end{equation}
and $\sigma_i$ is the $i$-th eigenvalue.


\end{document}